\documentclass[journal]{IEEEtran}
\usepackage{amsmath,amsfonts}
\usepackage{algorithmic}
\usepackage{algorithm}
\usepackage{array}
\usepackage[caption=false,font=normalsize,labelfont=sf,textfont=sf]{subfig}
\usepackage{textcomp}
\usepackage{stfloats}
\usepackage{url}
\usepackage{verbatim}
\usepackage{graphicx}
\usepackage{cite}
\usepackage{multirow}
\usepackage{booktabs}
\usepackage{pifont}

\usepackage{xcolor}
\usepackage{colortbl, booktabs}
\usepackage{hyperref}
\usepackage{soul, color}
\hypersetup{
    colorlinks=true,
    citecolor=blue,    
    linkcolor=blue,    
    urlcolor=blue,     
    filecolor=blue,   
    menucolor=blue,    
    runcolor=blue,    
    pdfborder={0 0 0}
}

\makeatletter
\def\@cite#1#2{\textcolor{blue}{[#1\if@tempswa , #2\fi]}}
\makeatother
\usepackage{cleveref}
\begin{document}

\title{ImagineNav++: Prompting Vision-Language Models as Embodied Navigator through Scene Imagination}

\author{Teng Wang$^{\dagger}$, Xinxin Zhao$^{\dagger}$, Wenzhe Cai, Changyin Sun

\thanks{Teng Wang, Xinxin Zhao, Changyin Sun are with the School of Automation, Southeast University, Nanjing 210096, China. (e-mail: \{wangteng, xinxin\_zhao, cysun\}@seu.edu.cn).  (\textit{Corresponding author : Teng Wang})}
\thanks{Wenzhe Cai is with Shanghai AI Lab, Shanghai, China. (e-mail: caiwenzhe@pjlab.org.cn).}
\thanks{$^{\dagger}$These authors contributed equally to this work.}}

\markboth{IEEE TRANSACTIONS ON PATTERN ANALYSIS AND MACHINE INTELLIGENCE}
{Shell \MakeLowercase{\textit{et al.}}: A Sample Article Using IEEEtran.cls for IEEE Journals}

\maketitle

\begin{abstract}
Visual navigation is a fundamental capability for autonomous home-assistance robots, enabling the execution of long-horizon tasks such as object search. While recent methods have leveraged Large Language Models (LLMs) to incorporate commonsense reasoning and improve exploration efficiency, their planning processes remain constrained by textual representations, which cannot adequately capture spatial occupancy or scene geometry--critical factors for informed navigation decisions. In this work, we explore whether Vision-Language Models (VLMs) can achieve mapless visual navigation using only onboard RGB/RGB-D streams, unlocking their potential for spatial perception and planning. We achieve this by developing the imagination-powered navigation framework \textit{ImagineNav++}, which imagines the future observation images at valuable robot views and translates the complex navigation planning process into a rather simple best-view image selection problem for VLMs. Specifically, we first introduce a \textit{future-view imagination} module, which distills human navigation preferences to generate semantically meaningful candidate viewpoints with high exploration potential. These imagined future views then serve as visual prompts for the VLM to identify the most informative viewpoint. To maintain spatial consistency, we develop a \textit{selective foveation memory} mechanism, which hierarchically integrates keyframe observations through a sparse-to-dense framework, thereby constructing a compact yet comprehensive memory for long-term spatial reasoning. This integrated approach effectively transforms the challenging goal-oriented navigation problem into a series of tractable point-goal navigation tasks. Extensive experiments on open-vocabulary object and instance navigation benchmarks demonstrate that our ImagineNav++ achieves SOTA performance in mapless setting, even surpassing most cumbersome map-based methods, revealing the importance of scene imagination and scene memory in VLM-based spatial reasoning. 

\end{abstract}


\section{Introduction}
\IEEEPARstart{A}{} useful home-assistant robot should be able to search for different kinds of objects without access to precise 3D spatial coordinates. This capability becomes particularly essential given the dynamic nature of household environments, where new objects are continuously introduced through regular human acquisition patterns. Consequently, the robot's navigation system requires open-set visual grounding and searching functionality that extends beyond predefined categorical constraints. In the research community, this fundamental challenge is formally characterized as the open-vocabulary goal-oriented visual navigation problem. Depending on the type of goal specification, several sub-tasks have been studied within this paradigm. Among these, we focus on two representative categories: Object-Goal Navigation (ObjectNav), in which the target is specified by a semantic category~\cite{topiwala2018frontier, cai2023bridgingzeroshotobjectnavigation, chaplot2020object, ramrakhya2022habitat}, and Instance-Image-Goal Navigation (InsINav), where the goal is defined by a reference image of a specific object instance \cite{sun2024prioritizedsemanticlearningzeroshot, yin2025unigoaluniversalzeroshotgoaloriented}.

\begin{figure*}[ht]
\centering
\includegraphics[width=1.00\linewidth]{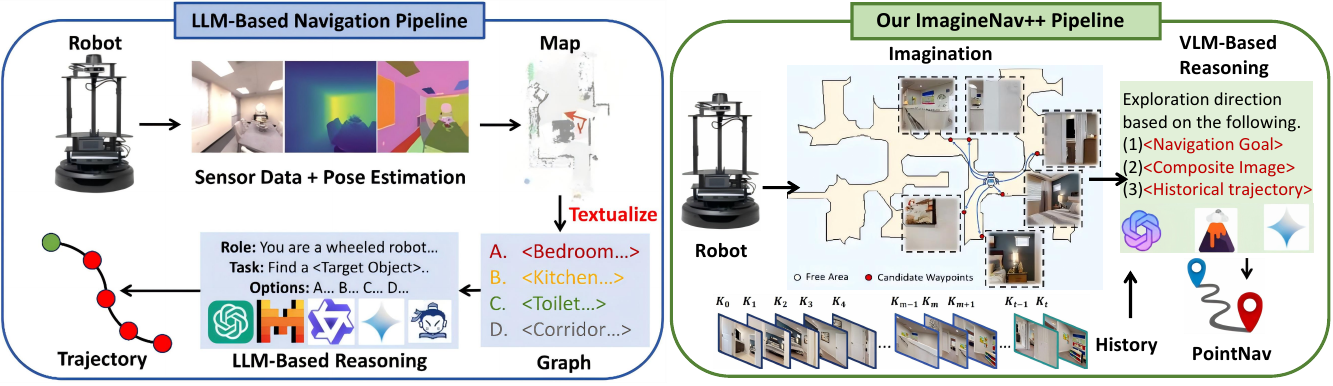}
\caption{The comparison between the conventional LLM-based navigation pipeline and our ImagineNav++ pipeline. The traditional LLM-based navigation framework, illustrated on the left, relies on intricate sensor data processing and pose estimation for map creation, followed by LLM-driven reasoning to decide the exploration direction. Instead, our ImagineNav++ directly decomposes the long-horizon object goal navigation task into a sequence of best-view image selection tasks for VLM, which avoids the latency and compounding error in the traditional cascaded methods.}
\label{fig1}
\vspace{-0.25cm}
\end{figure*}

Recent advances in foundation models, including vision models \cite{radford2021learning,he2022masked,zhou2022detecting,Cheng2024YOLOWorld,kirillov2023segment,wu2024general,liu2023grounding}, large language models (LLMs)~\cite{brown2020language,chowdhery2023palm,touvron2023llama,zhang2022opt}, and vision-language models (VLMs)~\cite{achiam2023gpt,team2023gemini,liu2024visual,chen2024internvl,Dai2023InstructBLIPTG,gao2023llama}, have enabled new possibilities for developing agents capable of open-vocabulary goal-oriented visual navigation. A widely adopted framework, as illustrated in Figure~\ref{fig1}, employs a modular approach to address this challenge, typically comprising four key components: (1) \textit{A real-time mapping and segmentation module} to construct a geometric and semantic representation of the robot's immediate environment through simultaneous localization and mapping (SLAM) combined with pixel-wise semantic segmentation~\cite{he2018maskrcnn, kirillov2023segment}; (2) \textit{A template-based translation module} to transform the structured semantic map into natural language descriptions, enabling compatibility with language models; (3) \textit{A LLM-based reasoning module} to generate step-by-step navigation plans in texts, incorporating task constraints and commonsense reasoning; (4) Finally, \textit{a path-planning module} to projects the LLM's output back onto the environmental map, computing optimal collision-free trajectories towards the goal using off-the-shelf motion planning algorithms. This pipeline effectively combines the strengths of visual perception, language understanding, and robotic control to achieve open-vocabulary navigation capabilities.

Although such pipelines have achieved significant success in recent years~\cite{Zhou2023ESCEW,kuang2024openfmnav,wu2024voronav,zhang2024trihelper,shah2023navigationlargelanguagemodels,Yu2023L3MVNLL,loo2025open}, these cascaded systems face several inherent limitations. First, both the depth camera and the robot localization module can suffer from perception error, especially for long-range depth estimation, and this can make the mapping process inaccurate. Second, the robotic system must maintain continuous real-time object detection and semantic segmentation to simultaneously enrich spatial mapping with semantic metadata and generate structured inputs for LLM-based reasoning. These computationally intensive operations introduce a critical performance bottleneck, substantially elevating the processing requirements of the robotic platforms. Third, while semantic information stored in the map can be readily represented in textual form--for instance, by listing observed object categories--such text-only prompts struggle to convey explicit geometric relationships and fine-grained object details. This limitation makes it difficult and ambiguous for LLMs to infer the best navigation plan.

In this work, we try to explore whether it is possible to circumvent the complicated and fragile \textit{mapping$\rightarrow$ translation$\rightarrow$ planning} framework, instead developing a visual navigation approach that operates directly on raw RGB/RGB-D observations using pre-trained VLMs. Our proposed \textit{\textbf{ImagineNav++}} framework seeks to maximize the capabilities of VLMs in multimodal scene understanding and spatial reasoning, effectively transforming VLMs into efficient embodied navigation agents. However, due to fundamental architectural constraints, current VLMs lack the capability for continuous 3D spatial reasoning~\cite{yamada2024evaluatingspatialunderstandinglarge}, rendering them unsuitable for direct generation of navigable 3D waypoints. To overcome this limitation, we propose a novel paradigm that reformulates visual navigation as an imagination-powered best-view image selection task, strategically harnessing VLMs' discriminative image analysis capabilities while circumventing their inherent limitations in geometric reasoning. Addressing the central challenge of generating well-founded candidate views for VLM-based selection, we develop the \textit{\textbf{Where2Imagine}} module, which distills human indoor navigation habits to generate future 3D navigation waypoints where a human might navigate based on the current visual observation. Such 3D navigation waypoints indicate relative poses with respect to the current frame and can be readily translated into new observation images using novel view synthesis (NVS) models~\cite{yu2024polyoculussimultaneousmultiviewimagebased,tseng2023consistentviewsynthesisposeguided,yu2023longtermphotometricconsistentnovel}. Afterwards, the VLMs only need to select the most relevant imagined observation with respect to the target object and drive the robot to follow the corresponding point-goal navigation trajectory. 

Furthermore, it is well-established that agent memory plays a crucial role in embodied navigation, allowing agents to accumulate, interpret, and leverage historical scene observations to inform future decisions~\cite{fukushima2022object,wang2023gridmm}. A major open challenge that arises in the context of VLM-based view selection is how to effectively encode the continuously expanding stream of observations to endow pre-trained VLMs with temporal reasoning capabilities. A common approach leverages VLMs to caption individual observations, subsequently aggregating these textual summaries over time~\cite{zhou2023navgpt,zhang2025mem2ego}. However, this abstraction process may inadvertently discard essential spatial and semantic information inherent in raw visual data, fundamentally limiting effective spatio-temporal reasoning. An alternative paradigm directly employs historical image sequences for end-to-end decision-making~\cite{zhang2024navid,zhang2024uni,tsai2023multimodal}. While architecturally simpler, these approaches often exhibit limited capacity to model long-term dependencies and show reduced robustness against perceptual noise in lengthy visual sequences. To overcome these limitations, we develop a \textit{\textbf{Selective Foveation Memory}}, which mimics human foveation mechanisms by hierarchically integrating keyframe observations through a sparse-to-dense framework--maintaining sparse keyframes for long-range structural context while preserving dense frames for proximate details, thereby efficiently capturing both global layout and local semantics. In particular, we introduce the innovative use of the pre-trained vision foundation model DINOv2~\cite{DINOv2} to semantically interpret historical observations, measure inter-frame similarity, and extract representative keyframes from extended observation sequences, effectively addressing the redundancy in long-term visual streams while preserving essential spatio-temporal structures for robust decision-making. 

The above pose-aware imagination-and-selection capability, integrated with the selective foveation memory allows the goal-oriented visual navigation task to be effectively decomposed into a sequence of point-goal sub-tasks, facilitating the creation of collision-free navigation trajectories. Experimental results on standard benchmarks demonstrate the superiority of our ImagineNav++ in open-vocabulary object/instance navigation. Our main contributions are highlighted as follows:

\begin{itemize}
    \item We develop a mapless navigation framework ImagineNav++. It leverages the imagination to generate image observations at potential future 3D waypoints as the VLMs' visual prompts, grounding the VLMs to become efficient navigation agents without any fine-tuning.
    \item We design a task-oriented model Where2Imagine to understand human navigation habits. This model is crucial to bridge the task-agnostic high-level VLM planners and the low-level navigation policies.
    \item We introduce a selective foveation memory mechanism that hierarchically integrates keyframe observations in a sparse-to-dense manner, thereby constructing a discriminative and compact scene representation for long-term spatial reasoning. 
    \item For zero-shot ObjectNav, our ImagineNav++ increases success rate by a large margin of 4.0\% and 23.5\% respectively on the complex HM3D~\cite{ramakrishnan2021hm3dc} and HSSD~\cite{khanna2023habitatsyntheticscenesdataset}, while achieving competitive result on Gibson~\cite{xia2018gibson}. For InsINav, our approach attains the highest SPL on HM3D. 
\end{itemize}

This paper is an extended version of our prior publication~\cite{zhao2025imaginenavpromptingvisionlanguagemodels} in ICLR 2025. The main differences from the conference version are listed as follows: 1) We introduce a selective foveation memory that maintains spatial consistency via structured trajectory reasoning, thereby mitigating prior decision biases through its hierarchical keyframe selection and organization. 2) We demonstrate the method's robust transferability by showing that it could seamlessly adapt to the InsINav task with only minimal architectural adjustments. 3) Our empirical findings are further  supported by extensive evaluation across multiple benchmarks, complemented by detailed ablation studies and hyperparameter analysis.

\section{Related work}
\subsection{Large Models for Robotic Planning}
Large-scale models pre-trained on extensive internet data have demonstrated formidable zero-shot reasoning capabilities in tasks such as planning~\cite{huang2022languagemodelszeroshotplanners}, code generation~\cite{liang2023codepolicieslanguagemodel,huang2023instruct2act}, and solving science questions~\cite{lewkowycz2022solvingquantitativereasoningproblems}. The in-context learning capability of LLMs allows them to be applied to robotic task planning. Some methods~\cite{liang2023codepolicieslanguagemodel,huang2023instruct2act,ahn2022icanisay} leverage LLMs to decompose tasks into subtasks, enhancing execution efficiency. Cap~\cite{liang2023codepolicieslanguagemodel} generates robotic policy code directly from example language commands, enabling autonomous control and task execution based on natural language instructions. Instruct2Act~\cite{huang2023instruct2act} combines LLM with foundational models (e.g., SAM and CLIP), reducing error rates in complex task execution, while SayCan~\cite{ahn2022icanisay} combines LLM task planning with the feasibility of physical skills using pre-trained value functions, generating actionable plans for robots. However, one limitation of LLMs is their difficulty in embedding the robot’s state directly into the planning process. To address this, many studies have turned to VLMs as alternatives. For instance, ViLA~\cite{lin2024vilapretrainingvisuallanguage} significantly improves performance on multimodal tasks without compromising text capabilities by systematically exploring VLM pretraining design choices. CoPa~\cite{huang2024copageneralroboticmanipulation} incorporates commonsense knowledge from VLMs, proposing a coarse-to-fine task-oriented grasping and task-aware motion planning approach. PIVOT~\cite{nasiriany2024pivotiterativevisualprompting} transforms tasks into iteratively optimized visual question-answering problems via a refinement process. Socratic model~\cite{zeng2022socraticmodelscomposingzeroshot} integrates multiple pretrained large models (e.g., VLMs, LLMs, and audio models) in a modular fashion to enable reasoning and task execution through language-based interaction. These methods employ a set-of-examples (SOE) approach to guide VLM selection. We propose a new decision-making paradigm based on imagined imagery, wherein decisions are made on imaginations, enabling more nuanced, context-aware interactions that better harness VLMs’ spatial perception capabilities.

\subsection{Open-Vocabulary Navigation: From Objects to Instances}
Open-vocabulary navigation requires agents to respond to instructions involving object categories or specific instances not encountered during training, placing higher demands on the model's generalization and grounding capabilities. In object goal navigation, early end-to-end methods attempted to integrate textual knowledge into navigation tasks by leveraging compact multimodal feature spaces such as CLIP~\cite{radford2021learning}, yet their performance remains limited~\cite{khandelwal2022simple,gadre2023cows,majumdar2022zson}. Modular approaches~\cite{huang2023visuallanguagemapsrobot,Zhou2023ESCEW,achiam2023gpt}, on the other hand, typically rely on sensors for localization and mapping, high-level planning, and low-level control. Their dependence on hardware precision and pre-computation constrains practical flexibility. Meanwhile, instance goal navigation presents greater challenges due to its requirement for distinguishing unique target instances. Representative works such as Goat~\cite{chang2024goatthing} achieve lifelong navigation by constructing an instance-aware semantic memory and integrating classical path planning. Mod-IIN~\cite{krantz2023navigatingobjectsspecifiedimages}, is specifically designed for this task, utilizing feature matching and map projection for goal re-localization. UniGoal~\cite{yin2025unigoaluniversalzeroshotgoaloriented} employs a unified graph representation and graph matching to handle diverse goal types. While these methods demonstrate strong performance on specific benchmarks, most still rely heavily on explicit environmental representations--such as maps or scene graphs--or extensive task-specific training. Our approach introduces an imagination-based, mapless navigation framework. This framework circumvents the need for extensive training by transforming the complex process of navigation planning into a selection problem based solely on RGB inputs.

\begin{figure*}[ht]
\centering
\includegraphics[width=1.00\linewidth]{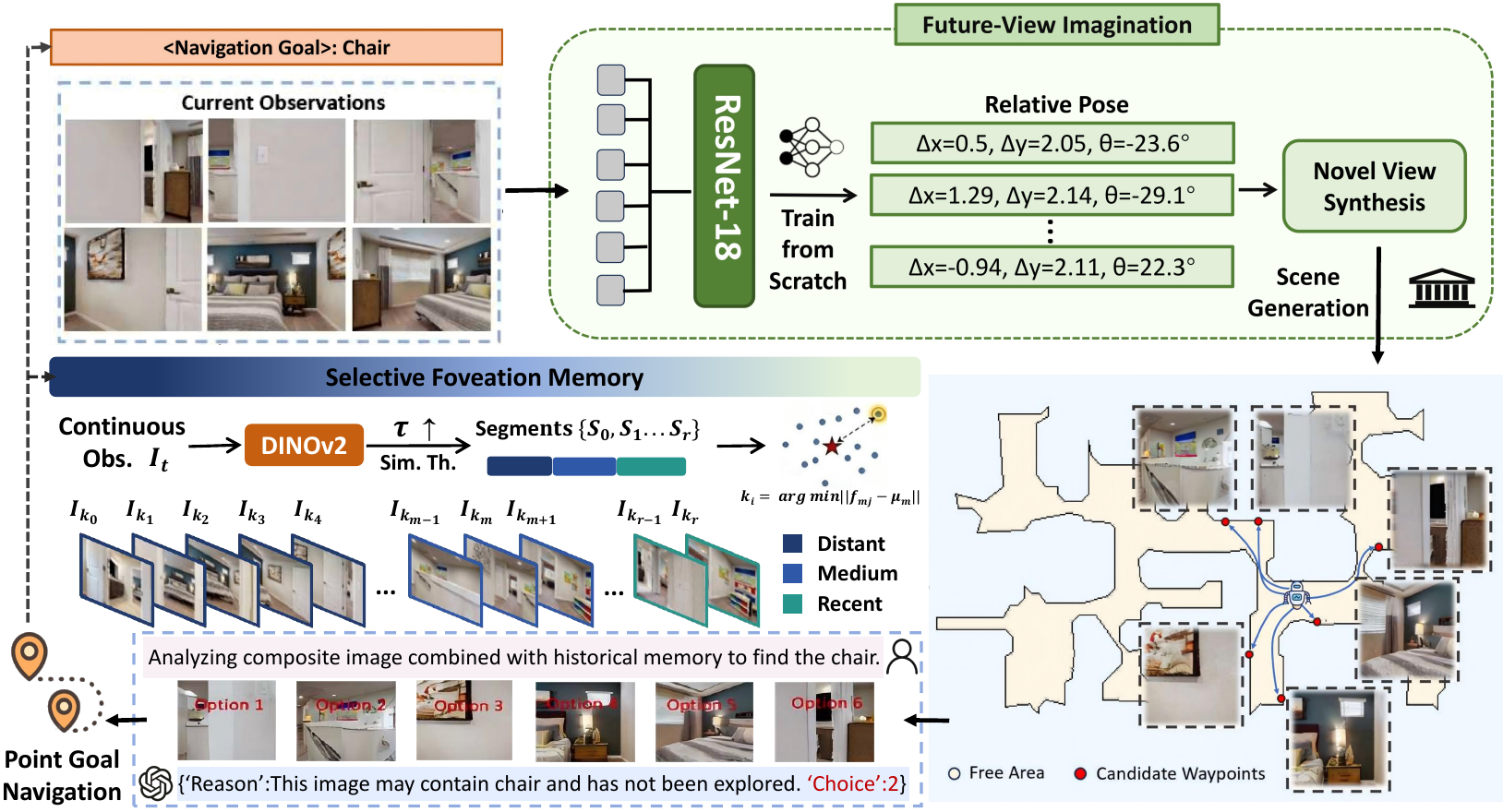}
\caption{The overall pipeline of our mapless, open-vocabulary navigation framework ImagineNav++. At each iteration, the agent captures a panoramic view of its surroundings. The Imagination Module then leverages the trained Where2Imagine module coupled with a novel view synthesis model to generate novel scene views. Guided by structured prompts, the VLM engages in target-oriented inference by integrating historical selective foveation memory with imagined future waypoint observations. Subsequently, the system executes the PointNav policy to determine the next navigational action. The above imagination, reasoning and planning procedure iterates until the target is reached.}
\label{fig2}
\vspace{-0.25cm}
\end{figure*}

\subsection{Imagination in Embodied Navigation}
Recent methods~\cite{zhai2022peanutpredictingnavigatingunseen,ramakrishnan2022poni,zhu2022navigatingobjectsunseenenvironments} have adopted supervised learning to infer potential target locations, addressing the `Where to look?' challenge in navigation. These approaches typically predict either the absolute coordinates~\cite{zhai2022peanutpredictingnavigatingunseen} of the target, the shortest distance to target~\cite{ramakrishnan2022poni}, or the nearest boundary~\cite{zhu2022navigatingobjectsunseenenvironments} based on local maps. Another line of research~\cite{ramakrishnan2020occupancyanticipationefficientexploration,georgakis2021learning,9560925,Zhang_2024_CVPR} has focused on enhancing the prediction of unobserved regions through diverse methodologies. For instance,~\cite{ramakrishnan2020occupancyanticipationefficientexploration} introduced occupancy anticipation, where the agent infers an occupancy map based on RGB-D inputs. The L2M framework was introduced~\cite{georgakis2021learning}, consisting of a two-stage segmentation model that generates a semantic map beyond the agent’s field of view and selects long-term goals based on the uncertainty of predictions. SSCNav~\cite{9560925} leverages semantic scene completion and confidence maps to infer the environment and guide navigation decisions. A self-supervised generative map (SGM) is proposed~\cite{Zhang_2024_CVPR}, which employs self-supervised approach to continually generate unobserved regions in the local map and predict the target's location. These methods primarily predict unobserved regions in top-down maps derived from egocentric RGB-D projections. By contrast, our ImagineNav++ performs future-scene imagination directly in RGB space. We employ a compact model aligned with human navigation behavior to generate novel viewpoint locations, which are then translated into corresponding visual observations through a diffusion-based view synthesis process.

\subsection{Memory Mechanism in Embodied Navigation}
\vspace{-0.1cm}
Memory plays a crucial role in visual navigation by improving an agent’s perceptual consistency and long-term reasoning capabilities in unfamiliar environments~\cite{fukushima2022object,wang2023gridmm}. Initially, LLMs were predominantly applied in embodied navigation systems to support high-level decision-making. To enable LLMs to process perceptual inputs, earlier approaches~\cite{zhou2023navgpt, zhang2025mem2ego} generate descriptive captions of individual observations and aggregate these textual summaries over time. Subsequent methods typically construct a globally consistent memory map from visual observations, infer potential navigational waypoints through structured spatial reasoning, and encode these into natural language representations suitable for LLMs~\cite{shah2023navigationlargelanguagemodels, wu2024voronav, kuang2024openfmnav, Zhou2023ESCEW}, such as the semantic score map in OpenFMNav~\cite{kuang2024openfmnav} and the reduced Voronoi graph in VoroNav~\cite{wu2024voronav}. 
The inherent loss of geometric information in language representations has motivated the development of hybrid memory approaches that maintain structured environmental state values (e.g., target distance and exploration reward). For instance, VLMap~\cite{huang2023visuallanguagemapsrobot} quantifies navigation value via pixel-level feature-text similarity, while VLFM~\cite{yokoyama2024vlfm} projects image-text similarity into dynamic value maps using BLIP-2~\cite{li2023blip}. While effectively mitigating memory forgetting, these methods require extensive precomputation and environment-specific training, limiting their scalability to unseen environments.
With recent advances in multimodal learning, VLMs have been increasingly integrated into visual navigation systems to improve scene interpretation and decision-making~\cite{goetting2024end,tang2025openin,gao2025octonav,zhang2024navid,zhang2024uni,tsai2023multimodal}.
VLMNav~\cite{goetting2024end} constructs dynamic traversability masks in a VQA framework. Concurrently, mapless navigation paradigms have emerged, bypassing explicit mapping to leverage multimodal observations. Representative methods include NaVid~\cite{zhang2024navid}, which encodes monocular RGB streams into spatio-temporal representations using special tokens to differentiate historical and current observations; its extension Uni-NaVid~\cite{zhang2024uni} incorporating hierarchical token compression for efficient long-context processing; and the GRU-based history aggregation approach of~\cite{tsai2023multimodal}, which models temporal observations to predict collision-aware action distributions. In line with these mapless navigation paradigms, our method also utilizes historical visual observations as direct input to VLMs. However, a key distinction of our approach lies in its use of \textit{visually salient keyframes}, which compress perceptual history, enhance temporal coherence, and emphasize task-critical visual information, leading to more robust and efficient navigation performance.

\vspace{-0.15cm}
\section{Methodology}
\subsection{Problem Formulation and Overall Framework}
The open-vocabulary goal-oriented navigation task requires an agent to locate an arbitrary target object instance in novel environments, without prior exposure to the object category during training. At the beginning of each episode, the agent is initialized at a random pose with no prior knowledge of the scene layout. The goal is to find a target object \(g_i\), which can belong to any category in an open-vocabulary setting. In ObjectNav, the target is specified by semantic category, whereas in InsINav, it is defined by a reference image of a particular object instance. At each time step \(t\), the agent receives an egocentric panorama view \(I_t\), divided into 6 separate views, each represented by an RGB image $I_{t, i}$ accompanied by its depth map $D_{t, i}$. The agent operates over a discrete low-level action space comprising: \{\texttt{Stop}, \texttt{MoveAhead}, \texttt{TurnLeft}, \texttt{TurnRight}, \texttt{LookUp}, \texttt{LookDown}\}. The \texttt{MoveAhead} action moves the agent forward by 25 cm, while the rotational actions \texttt{TurnLeft} and \texttt{TurnRight} rotate the agent by $30$ degrees. The task is considered successful if the agent reaches the target object with a geodesic distance smaller than a defined threshold (e.g., 1m) and executes the \texttt{Stop} command within a fixed number of steps. Each episode is limited to a maximum of 500 steps. 

An overview of our proposed imagination-based open-vocabulary visual navigation framework, dubbed \textbf{ImagineNav++}, is presented in Figure~\ref{fig2}. The agent first utilizes the \textit{Where2Imagine} module to generate candidate waypoints for imagination based on the current visual observation. A novel view synthesis (NVS) model then renders imagined visual observations corresponding to these candidate locations.  By leveraging the agent’s scene memory, the multimodal large-scale model evaluates the synthesized images, each annotated with option labels, to assess both spatial structure and semantic coherence, enabling context-aware and efficient selection of exploration directions. Specifically, a vision-language model (VLM) is prompted to reason over the six-view imagined future observations and select the optimal waypoint. The agent then employs a low-level point navigation policy to reach the chosen sub-goal. Once the subgoal is deemed reached, the memory is updated by recursively incorporating the new observations into the hierarchical memory structure. This procedure iterates recursively-each new observation serves as input to subsequent imagination, reasoning, and navigation steps-until an instance of the target object is successfully identified. Through this mechanism, the goal-oriented visual navigation task is decomposed into a sequence of manageable point-goal navigation sub-tasks. A key advantage of our proposed ImagineNav++ is its training-free reasoning and planning pipeline, which requires no object-specific data, making it inherently open-vocabulary and capable of zero-shot generalization to unseen semantic targets. Section~\ref{subsec:imagination} introduces the imagination module, Section~\ref{subsec:memory} describes the memory module, Section~\ref{subsec:planner} details the VLM-based planner, and Section~\ref{subsec:controller} describes the point navigation policy.
\vspace{-0.15cm}
\subsection{Future-View Imagination}
\label{subsec:imagination}
To better leverage the spatial perception and reasoning capabilities of VLMs for open-vocabulary visual navigation in unknown environments, we propose a future-view imagination model, which is composed of a \textit{Where2Imagine} module followed by a novel view synthesis (NVS) model. As illustrated in Figure~\ref{fig2}, the \textit{Where2Imagine} predicts the relative pose (\(\Delta x,\Delta y, \Delta \theta\)) of a potential next waypoint from the current RGB observation, where \(\Delta x\), \(\Delta y\) and $\Delta \theta$ denote lateral displacement, longitudinal displacement, and viewing angle change, respectively. This predicted pose is subsequently fed into the NVS module to generate a corresponding imagined observation. Recent years have witnessed remarkable progress in NVS, with advanced methods ranging from few-shot 3D rendering techniques~\cite{sargent2024zeronvs, wimbauerbehind, cao2023scenerf} to generative models, particularly diffusion models~\cite{yu2024polyoculussimultaneousmultiviewimagebased, tseng2023consistentviewsynthesisposeguided, yu2023longtermphotometricconsistentnovel}, which have demonstrated powerful image synthesis capabilities across diverse domains. In this work, we adopt a pre-trained diffusion model ''Polyoculus''~\cite{yu2024polyoculussimultaneousmultiviewimagebased} for future-view imagination owing to its capacity to generate perceptually consistent and high-fidelity novel views from a single RGB image and a relative pose.
\vspace{-0.1cm}
\begin{figure}[!ht]
\centering  
\includegraphics[width=0.5\textwidth]{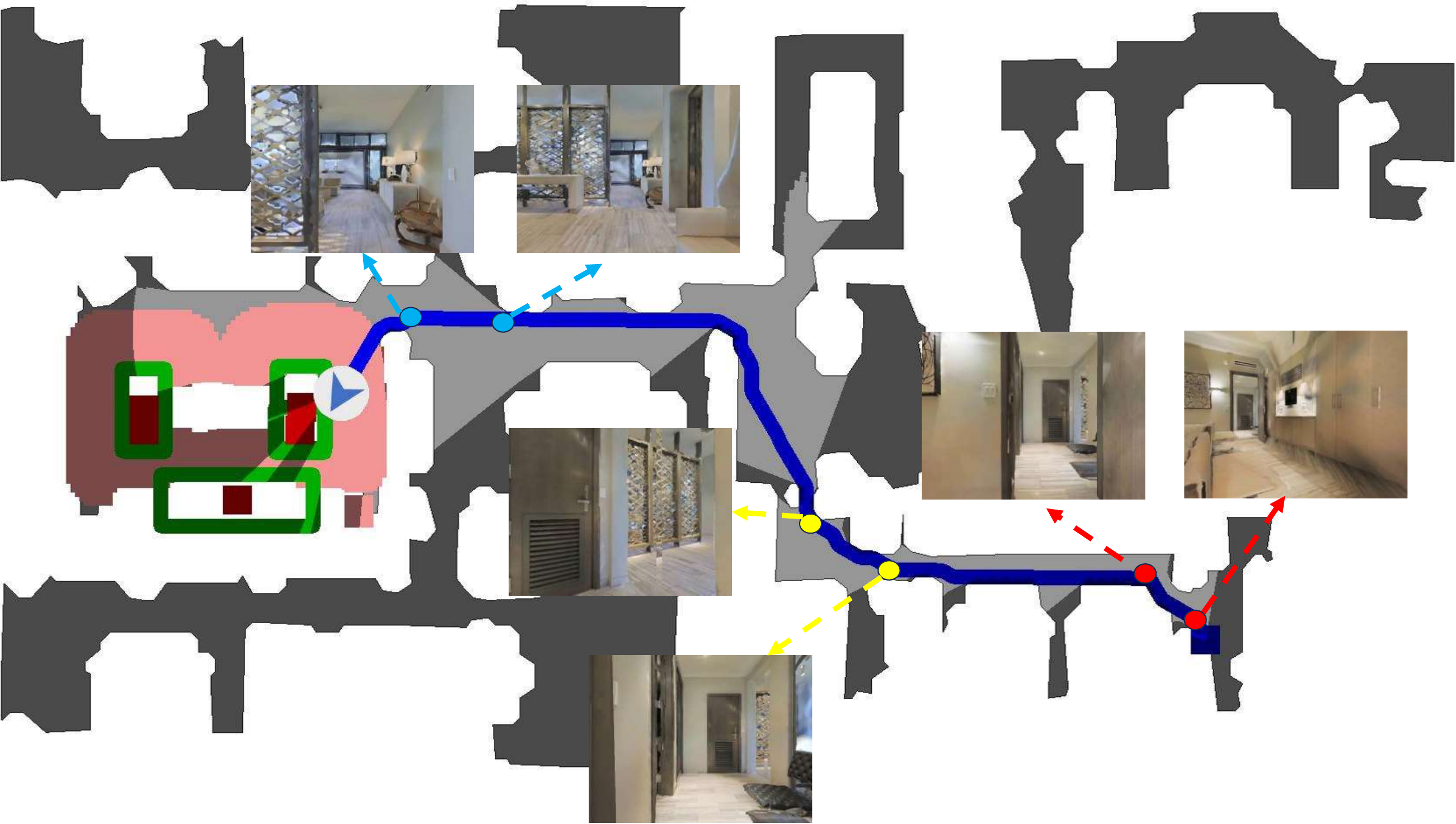}
\caption{Visualization of human demonstration trajectories on the MP3D dataset in the Habitat-Web project. The trajectories reveal a consistent tendency for humans to prioritize directions toward semantically meaningful cues (e.g., doors) that structurally facilitate efficient exploration.}
\label{fig:demonstration}
\end{figure}
\vspace{-0.1cm}

Specifically, to equip the \textit{Where2Imagine} module with human-like spatial intuition for waypoint prediction, we train a ResNet-18 model~\cite{he2016resent} from scratch using human demonstration data sourced from the Habitat-Web project~\cite{ramrakhya2022habitat}. The data comprises 80k ObjectNav and 12k Pick\&Place trajectories collected via a virtual teleoperation system on Amazon Mechanical Turk, capturing natural human navigation behaviors in indoor environments. A key insight derived from these demonstrations is that humans consistently favor directions toward semantically meaningful structures (e.g., doors) that facilitate exploration, as illustrated in Figure~\ref{fig:demonstration}. By learning from these demonstrations, the model acquires semantically-informed navigational preferences, enabling it to predict waypoints that are both geometrically plausible and semantically meaningful, thereby supporting efficient and contextually-aware navigation. We reformulated the human demonstration trajectories into a paired dataset $\{(O_t,P_{t+T})\}$, where $P_{t+T} = (\Delta x,\Delta y, \Delta \theta)$ denotes the ground-truth relative waypoint pose with respect to the agent’s current frame $O_t$. Here, $T$ defines a predictive temporal horizon that captures how humans anticipate and navigate through semantically meaningful spatial scales. To enhance learning efficacy, we apply a depth-based filter to exclude observations with limited semantic content (e.g., close-up views of blank walls):
\begin{equation}
    O_t~\text{is retained only if}~ d(O_t) \geq 0.3
\end{equation}
where $d(O_t)$ denotes the average depth of the visual observation $O_t$. Furthermore, considering the inherent constraints of diffusion-based NVS models in handling extreme viewpoint changes (e.g., 120°, 180°, 240°), we restricted the angular deviation $\Delta \theta$ in the training set to:
\begin{equation}
    \lvert \Delta \theta \rvert \leq 30^{\circ},
\end{equation}
ensuring synthesis quality and perceptual consistency. This curation procedure enhances training stability and promotes the generation of plausible, human-like waypoints through the \textit{Where2Imagine} module. Finally, The model is trained end-to-end using a mean squared error~(MSE) loss between the predicted and ground-truth waypoint poses:
\begin{equation}
    \mathcal{L}_{\text{waypoint}} = \frac{1}{B} \sum_{b=1}^{B} \left\| P^{(b)}_{t+T} - \hat{P}^{(b)}_{t+T} \right\|^2_2
\end{equation}
where $B$ denotes the batch size and $\left\|\cdot\right\|_2$ refers to $L_2$-norm. 

During inference, at navigation time step $t$, the model predicts the relative waypoint pose from each of the six current views $I_{t, i}, i\in\{1,2,\cdots, 6\}$, as follows:
\begin{equation}
\hat{P}_{t+T, i} = [\hat{\Delta} x_{t, i},\hat{\Delta} y_{t, i},\hat{\Delta}\theta_{t,i} ] = \text {ResNet-18}(I_{t,i}),
\end{equation}
where $\hat{P}_{t+T,i}$ denotes the predicted relative displacement and heading angle of the waypoint at a future time step $t+T$. The set $\{(I_{t, i}, \hat{P}_{t+T, i})\}_{i=1}^6$ is then passed to  the diffusion model to generate imagined waypoint observations $\{M_{t,i}\}_{i=1}^{6}$. 

\subsection{Selective Foveation Memory}
\label{subsec:memory}
In complex indoor navigation scenarios, keyframes serve as compact yet informative representations that capture salient spatial and contextual information across temporal sequences. They reduce data redundancy and preserve essential perceptual cues, thereby enhancing memory efficiency while supporting long-term temporal reasoning. However, reliably extracting such keyframes from observational data remains challenging, particularly in identifying semantically meaningful segments within extended visual streams. Traditional keyframe extraction methods often rely on hand-crafted features such as color histograms~\cite{VSUMM2011, Ejaz2012,Wu2017} for clustering, which struggles to capture deep semantic similarities between frames. In contrast, modern deep learning approaches can model complex spatio-temporal dependencies but are typically heavily supervised, requiring extensive manual annotations~\cite{Zhao2018, Rochan2018, Elfeki2019, Jiang2022}. This motivates our exploration of DINOv2~\cite{DINOv2} for keyframe extraction in a zero-shot manner, where selection is driven primarily by inter-frame visual similarity. DINOv2's self-supervised pre-training on a large-scale and diverse image corpus enables it to learn unified visual representations that inherently encode holistic scene structure. Consequently, the model exhibits a pronounced sensitivity to scene-level variations, making it particularly suitable for identifying semantically representative keyframes without task-specific training. 

Specifically, given a historical observation sequence $\boldsymbol{O} = \{I_1, I_2, \dots, I_t\}$ up to time $t$, we first encode each observation using DINOv2 to extract discriminative feature embeddings $\boldsymbol{F} =\{f_1, f_2, \dots, f_t\}$. To construct a compact yet informative memory, we employ a semantic similarity-based keyframe selection strategy. Specifically, moving backward through the entire sequence, we measure the cosine similarity between consecutive frame features as:
\begin{equation}
s_{i,i+1} = \frac{f_i \cdot f_{i+1}}{|f_i| |f_{i+1}|}
\end{equation}
Adjacent frames $I_i$ and $I_{i+1}$ are grouped into the same semantic segment if $s_{i,i+1} > \tau$; otherwise, $I_{i}$ initiates a new segment. Here, $\tau$ is a time-varying threshold controlling segmentation granularity: a higher $\tau$ leads to denser memory, while a lower $\tau$ encourages sparser retention. This aggregation procedure ultimately yields $M$ semantic segments $\{\mathcal{S}_1, \mathcal{S}_2, \cdots, \mathcal{S}_M\}$, from each of which we select the frame closest to the segment’s average feature as the keyframe. Formally, for each segment $\mathcal{S}_m$ containing frames $\{I_{m_1}, I_{m_2}, \dots, I_{m_{N_{m}}}\}$, we compute its feature centroid as:
\begin{equation}
\mu_m = \frac{1}{N_{m}}\sum_{j=1}^{N_{m}} f_{m_j},
\end{equation}
The selected keyframe $I_{k_m}$ from $\mathcal{S}_m$ is then determined by:
\begin{equation}
k_m = \arg\min_{j \in {1,\dots,N_m}} || f_{m_j} - \mu_m ||.
\label{eq:keyframe_selection}
\end{equation}
The resulting set of representative keyframes across all segments is denoted as $\mathcal{M} = \{I_{k_1}, I_{k_2}, \cdots, I_{k_M}\}$.

Furthermore, in embodied visual navigation, visual observations from different time steps actually serve distinct purposes. To be precise, recent observations capture fine-grained local details, while far historical observations establish a coherent global context. Building upon these insights, we partition the observation history $\boldsymbol{O}$ up to the current time step $t$, into three temporal segments, including distant-term $\boldsymbol{T}_d$, medium-term $\boldsymbol{T}_m$, and recent-term $\boldsymbol{T}_r$, using a keyframe-count-driven criterion. We assign decreasing thresholds $\tau_d < \tau_m < \tau_r$ to these segments, inducing monotonically increasing keyframe density from distant to recent segments. Specifically, starting from the current frame $I_t$ and moving backward:
\begin{itemize}
\item Recent-term memory $\mathcal{M}_r$ is formed by sequentially selecting and adding keyframes using similarity threshold $\tau_r$ until $N_r$ keyframes are selected.

\item Medium-term memory $\mathcal{M}_m$ then continues backward until $N_m$ keyframes are collected.

\item Distant-term memory $\mathcal{M}_d$ includes all other keyframes extracted from all prior observation frames via semantic segmentation using similarity threshold $\tau_d$.
\end{itemize}
This design emulates a foveated memory mechanism, emphasizing recent high-detail contexts while preserving structural coherence over the long term. The resulting hierarchical memory $\mathcal{M} = \{\mathcal{M}_r, \mathcal{M}_m, \mathcal{M}_d\}$ provides structured spatiotemporal context for VLMs, balancing real-time detail with long-term coherence.

\textbf{Efficient Variant for Real-Time Deployment}. Computing global feature similarities over the entire observation history incurs $\mathcal{O}(N)$ complexity, which is prohibitive for lifelong navigation. To address this, we propose an \textit{online, incremental-update} variant of the Selective Foveation Memory. Observations are processed in a streaming manner: each new frame $I_t$ is compared only with the last frame of the \textit{currently active segment}. If their similarity exceeds $\tau_r$, $I_t$ is absorbed; otherwise, the segment is terminated and compressed by extracting a representative keyframe using Eq.~\ref{eq:keyframe_selection} and storing it in a short-term buffer, along with the partitioning similarity for subsequent hierarchical grouping. To strictly bound memory usage, we further introduce a hierarchical capacity constraint. When the short-term buffer reaches capacity, the oldest keyframes are transferred to the medium-term memory $\mathcal{M}_m$. To preserve $\mathcal{O}(1)$ update complexity, medium- and long-term updates reuse the recorded similarities instead of recomputing them, grouping adjacent keyframes via $\tau_m$ and $\tau_d$ and extracting higher-level representatives (“keyframes of keyframes”). This hierarchical compression mechanism prevents out-of-memory (OOM) while preserving the global spatiotemporal structure.

\subsection{High-level Planning}
\label{subsec:planner}
The high-level planning module leverages the spatial awareness and commonsense reasoning capabilities of the VLM to select the direction most conducive to locating the navigation target. We employ GPT-4o-mini~\cite{openai2024gpt4omini} as the high-level planner due to its favorable balance between reasoning performance and practical efficiency.  To assist GPT-4o-mini in decision making, we designed a simple prompt template, requiring the VLM to summarize its choice in a JSON format containing \{ \text{`Reason'}, \text{`Choice'} \}. This format allows for a clear understanding of the VLM's reasoning process. As illustrated in Figure~\ref{fig2}, the VLM is provided with synthesized observations of potential future waypoints, historical memory, and the navigation goal. Guided by the hierarchical prompt, it analyzes the semantic content of each view to select the optimal exploration direction, returning its decision in a structured format. The complete prompt can be found in section~\ref{experiment}. This integration of imagined future observations and historical keyframes as visual prompts equips our ImagineNav++ with significantly enhanced spatial reasoning and long-term decision-making capabilities. First, VLMs are more skilled at handling multiple-choice decision tasks compared to 3D geometry question answering (i.e., directly inferring the 3D coordinates of next waypoints). Furthermore, the introduction of the imagined future observations augment the VLM's scene understanding by providing rich contextual information about distant or visually unclear objects, while the historical keyframes maintain a coherent representation of the environment’s state over time. This complementary mechanism effectively mitigates perceptual uncertainty and facilitates robust long-horizon reasoning. The proposed imagination-guided reasoning procedure--integrating waypoint imagination and high-level planning--operates on a periodic cycle: it is triggered once every $T$ steps, proposing a new sub-goal only after the low-level controller has successfully completed navigation to the current target. This design ensures an effective balance between computational efficiency and robust long-horizon navigation performance.

\subsection{Low-level Controller}
\label{subsec:controller}
Following waypoint selection by the high-level planner, the low-level controller executes a Point Goal Navigation (PointNav) strategy to reach each designated target. In contrast to ObjectNav, which depends on semantic cues from the environment, PointNav operates purely on spatial perception, using relative displacement commands (\(\Delta x\), \(\Delta y\)) without requiring semantic understanding. Multiple established methods exist for PointNav implementation~\cite{yang2023iplanner, roth2024viplanner, wijmans2022ver, liang2024mtg}. In our framework, we adopt Variable Experience Rollout (VER~\cite{wijmans2022ver}) as the underlying policy for action selection at each navigation step. VER combines the advantages of synchronous and asynchronous reinforcement learning paradigms, leading to improved training efficiency and sample utilization. As a result, the agent exhibits enhanced adaptability and generalization performance in novel and complex environments.

\begin{table*}[!t]
    \centering
    \begin{minipage}{\textwidth}
        \centering
        \caption{\textbf{ImagineNav++: Comparison with previous work on object-goal navigation.} The Where2Imagine module with $T$=11, utilizing ResNet-18 trained from scratch and GPT-4o-mini as the VLM, was evaluated over 200 epochs on the Gibson, HM3D, and HSSD datasets. ImagineNav++ uses NVS model to generate novel view images, while ImagineNav++(Oracle) uses real images of the candidate points to facilitate spatial reasoning.}
        \label{table1}
        {\small
        \resizebox{0.90\textwidth}{!}{
            \begin{tabular}{cccccccccc}
                \toprule
                \multirow{2}{*}{\textbf{Method}}  & \multirow{2}{*}{\textbf{Open-Vocabulary}} & \multirow{2}{*}{\textbf{Mapless}} & \multicolumn{2}{c}{\textbf{Gibson}} & \multicolumn{2}{c}{\textbf{HM3D}} & \multicolumn{2}{c}{\textbf{HSSD}} \\
                \cmidrule(lr){4-5}  \cmidrule(lr){6-7}  \cmidrule(lr){8-9} 
                & & & SR $\uparrow$ & SPL $\uparrow $ & SR $\uparrow$ & SPL $\uparrow$ & SR $\uparrow$ & SPL $\uparrow$\\
                \midrule
                FBE~\cite{topiwala2018frontier} & \ding{55} & \ding{55} & 64.3 & 28.3 & 33.7 & 15.3 & 36.0 & 17.7 \\
                SemExp~\cite{chaplot2020object} & \ding{55} & \ding{55} & 71.7 & 39.6 & 37.9 & 18.8 & - & -  \\
                Habitat-Web~\cite{ramrakhya2022habitat} & \ding{55} & \ding{51} & - & - & 41.5 & 16.0 & - & - \\
                OVRL~\cite{yadav2023offline} & \ding{55} & \ding{51} & - & - & 62.0 & 26.8 & - & - \\
                ESC~\cite{Zhou2023ESCEW} & \ding{51} & \ding{55} & - & - & 39.2 & 22.3 & - & - \\
                VoroNav~\cite{wu2024voronav} & \ding{51} & \ding{55} & - & - & 42.0 & 26.0 & 41.0 & 23.2 \\
                VLFM~\cite{yokoyama2024vlfm} & \ding{51} & \ding{55} & 84.0 & 52.2 & 52.5 & 30.4 & - & -  \\
                Goat~\cite{chang2024goatthing} & \ding{51} & \ding{55} & - & - & 50.6 & 24.1 & - & - \\
                SG-Nav~\cite{yin2024sgnavonline3dscene} & \ding{51} & \ding{55} & - & - & 54.0 & 24.9 & - & - \\
                UniGoal~\cite{yin2025unigoaluniversalzeroshotgoaloriented} & \ding{51} & \ding{55} & - & - & 54.5 & 25.1 & - & -  \\
                \midrule
                ZSON~\cite{majumdar2022zson} & \ding{51} & \ding{51} & 31.3 & 12.0 & 25.5 & 12.6 & - & - \\
                PixNav~\cite{cai2023bridgingzeroshotobjectnavigation} & \ding{51} & \ding{51} & -& - & 37.9 & 20.5 & - & - \\
                PSL~\cite{sun2024prioritizedsemanticlearningzeroshot} & \ding{51} & \ding{51} &- &- & 42.4 & 19.2 & - & - \\
                ImagineNav~\cite{zhao2025imaginenavpromptingvisionlanguagemodels} & \ding{51} & \ding{51} &- &- & 53.0 & 23.8 & 51.0 & 24.9 \\
                
                \rowcolor{gray!30}\textbf{ImagineNav++} & \ding{51} & \ding{51} & 72.4 & 42.8 & 58.5 & 26.6 & 64.5 & 27.9  \\
                 \rowcolor{gray!30}\textbf{ImagineNav++(Oracle)} & \ding{51} & \ding{51} & 77.1 & 53.2 & 62.5 & 32.8 & 67.5 & 30.3  \\
                \bottomrule
            \end{tabular}
        }}
    \end{minipage}
    \vspace{-0.25cm}
\end{table*}

\section{Experiment}
\label{experiment}
\subsection{Experimental Setup and Evaluation Metrics}
We evaluate the effectiveness and navigation efficiency of our proposed ImagineNav++ using the Habitat v3.0 simulator~\cite{puig2023habitat} on both Object-Goal Navigation (ObjectNav) and Instance-Image-Goal Navigation (InsINav). For ObjectNav, we conduct experiments on three widely adopted datasets: Gibson~\cite{xia2018gibson}, HM3D~\cite{ramakrishnan2021hm3dc} and HSSD~\cite{khanna2023habitatsyntheticscenesdataset}.
Specifically, we employ the ObjectNav training and validation splits for Gibson, as introduced in SemExp~\cite{chaplot2020object}, covering 27 scenes and 6 object categories. The HM3D dataset, employed in the Habitat 2022 ObjectNav Challenge, contains 2,000 validation episodes across 20 unique environments and 6 object categories.
The recently introduced HSSD dataset comprises 40 high-quality synthetic scenes with 1,200 validation episodes covering 6 object categories. For InsINav, we follow established protocols from prior works~\cite{sun2024prioritizedsemanticlearningzeroshot, yin2025unigoaluniversalzeroshotgoaloriented} to evaluate performance on the HM3D dataset~\cite{ramakrishnan2021hm3dc}. The experimental configuration adheres to the ObjectNav Challenge 2023 guidelines~\cite{habitatchallenge2023}. To support the data collection for the \textit{Where2Imagine} module, we utilize human demonstration trajectories from the MP3D dataset~\cite{Matterport3D} within the habitat-web project, using a camera height of 0.88 m and a horizontal field of view (HFOV) of 79°.

We evaluate visual navigation performance using two standard metrics: Success Rate (SR) and Success weighted by Path Length (SPL)~\cite{anderson2018evaluation}. SR is defined as the proportion of episodes in which the agent executes the STOP action within 1 meter of the target object. SPL is computed as:
\begin{equation}
\text{SPL}= \frac{1}{N} \sum_{i=1}^{N} S_i \left( \frac{\ell_i}{\max(p_i, \ell_i)} \right), 
\end{equation}
where \(S_i\) is a binary success indicator for episode \(i\), \(p_i\) denotes the path length traversed by the agent and \(\ell_i\) represents the shortest path (ground-truth) length.

\subsection{Baselines}

For ObjectNav, we conducted a comparative analysis of non-zero-shot and zero-shot object navigation methods to rigorously evaluate the superiority of our ImagineNav++ framework. FBE~\cite{topiwala2018frontier} pioneered a frontier-based exploration strategy, emphasizing the boundaries between explored and unexplored regions. SemExp~\cite{chaplot2020object} advanced this concept by implementing goal-directed semantic exploration through the construction of semantic maps. In addition, we examined non-mapping closed-set visual navigation baselines, including approaches based on imitation learning~\cite{ramrakhya2022habitat} and visual representation learning~\cite{yadav2023offline}. For zero-shot object navigation, we evaluate several mapping-based baselines ~\cite{Zhou2023ESCEW,wu2024voronav,yokoyama2024vlfm, yin2024sgnavonline3dscene, yin2025unigoaluniversalzeroshotgoaloriented, chang2024goatthing}. Among these, \cite{Zhou2023ESCEW,wu2024voronav,yokoyama2024vlfm, chang2024goatthing} maintains a semantic 2D map of the scene and leverages semantic knowledge to facilitate navigation towards target objects, while SG-Nav~\cite{yin2024sgnavonline3dscene} and UniGoal~\cite{yin2025unigoaluniversalzeroshotgoaloriented} represent the observed scene with 3D scene graph and leverage LLMs for explicit graph-based reasoning. Additionally, we also investigate RGB-based non-mapping navigation baselines, including ZSON~\cite{majumdar2022zson}, PSL\cite{sun2024prioritizedsemanticlearningzeroshot} and PixNav~\cite{cai2023bridgingzeroshotobjectnavigation}. Specifically, ZSON and PSL employ the CLIP model~\cite{radford2021learning} to embed both target images and object goals into a unified semantic space, enabling the training of a semantic goal-driven navigation agent; PixNav, on the other hand, utilizes pixel-level guidance provided by VLMs and LLMs to achieve pixel-accurate navigation.
\vspace{-0.03cm}

For InsINav, we compare with the supervised methods Krantz et al.~\cite{krantz2022instance}, OVRL-v2~\cite{yadav2023ovrl}, and zero-shot
methods Mod-IIN~\cite{krantz2023navigatingobjectsspecifiedimages}, Goat~\cite{chang2024goatthing}, UniGoal~\cite{yin2025unigoaluniversalzeroshotgoaloriented}, and PSL~\cite{sun2024prioritizedsemanticlearningzeroshot}. Krantz et al.~\cite{krantz2022instance} pioneered the InsINav task and established an end-to-end RL baseline based on proximal policy optimization (PPO~\cite{schulman2017proximal}) and variable experience rollout (VER~\cite{wijmans2022ver}). OVRL-v2~\cite{yadav2023ovrl} employs an architecture that integrates ViT with LSTM, achieving end-to-end navigation learning through self-supervised visual pre-training, while OVRL-v2-IIN is specifically fine-tuned for InsINav task using the protocol from OVRL-v2~\cite{yadav2023ovrl}. Among these zero-shot methods, Mod-IIN~\cite{krantz2023navigatingobjectsspecifiedimages} is specifically tailored for InsINav task, which re-identifies the goal instance in egocentric vision using feature-matching and localizes the goal instance by projecting matched features to a map. Goat~\cite{chang2024goatthing} enables lifelong navigation by constructing an instance-aware semantic memory, utilizing CLIP~\cite{radford2021learning} and SuperGlue~\cite{sarlin2020superglue} for multimodal goal matching, and integrating frontier exploration with classical path planning. In contrast, UniGoal~\cite{yin2025unigoaluniversalzeroshotgoaloriented} achieves universal navigation by aligning diverse goals and scene graphs into a shared space through a unified graph representation and leveraging graph matching to dynamically guide its exploration.

\vspace{-0.25cm}
\subsection{Numerical Evaluations on \texttt{ObjectNav} Task}
Table~\ref{table1} presents a comparative analysis of the proposed ImagineNav++ framework against prior state-of-the-art methods. On the Gibson dataset, our ImagineNav++ achieves state-of-the-art performance under the open-vocabulary and mapless setting. While VLFM~\cite{yokoyama2024vlfm} reports a higher SR (84.0\%) on Gibson, its reliance on real-time dense semantic-geometric mapping makes it vulnerable to cumulative localization drift, particularly in large-scale environments, which limits practical robustness and scalability. In contrast, ImagineNav++ attains a competitive 72.4\% SR without explicit mapping, demonstrating greater system robustness and deployment flexibility while retaining open-vocabulary understanding. This advantage is convincingly validated  on the larger and more complex HM3D and HSSD dataset. Specifically, on HM3D, our ImagineNav++ maintains leading performance among open-vocabulary methods, achieving a 58.5\% SR and a 26.6\% SPL, surpassing map-based VLFM~\cite{yokoyama2024vlfm} and graph-based Unigoal~\cite{yin2025unigoaluniversalzeroshotgoaloriented} by 6.0\% and 4.0\% respectively in success rate. Moreover, ImagineNav++ achieves the highest SR and SPL on the HSSD dataset. The above observations indicate that our ImagineNav++ demonstrates outstanding navigation performance across various settings, while maintaining low storage and computational complexities. Furthermore, since the pretrained NVS is directly employed without finetunned on the Gibson, HM3D and HSSD datasets, we see a disparity between the quality of images generated by the NVS model and real images, limiting the capability of our model to some extent. To explore the upper limits of our framework, we instead use real panoramic images-specifically, the observation at the pose predicted by the Where2Imagine module-as visual prompts for the VLM model. Notably, both the success rate and SPL exhibit obvious improvements, obtaining 77.1\%, 62.5\% and 67.5\% at SR respectively on Gibson, H3MD and HSSD benchmarks, which further demonstrates the superiority of our imagination-based navigation framework.

\begin{table*}[!t]
    \centering
    \begin{minipage}{\textwidth}
        \centering
        \caption{\textbf{ImagineNav++: Comparison with previous work on instance-image-goal navigation.} The Where2Imagine module with $T$=11, utilizing ResNet-18 trained from scratch and GPT-4o-mini as the VLM, was evaluated over 200 epochs on the HM3D dataset.}
        \vspace{-0.1cm}
        \label{table:instance}
        {\small
        \resizebox{0.75\textwidth}{!}{
            \begin{tabular}{ccccccc}
                \toprule
                \multirow{2}{*}{\textbf{Method}}  & \multirow{2}{*}{\textbf{Open-Vocabulary}} & \multirow{2}{*}{\textbf{Mapless}} & \multirow{2}{*}{\textbf{Universal}} & \multicolumn{2}{c}{\textbf{HM3D}} \\
                \cmidrule(lr){5-6} 
                & & & & SR $\uparrow$ & SPL $\uparrow $\\
                \midrule
                Krantz et al.~\cite{krantz2022instance} & \ding{51} & \ding{55} & \ding{55} & 8.3 & 3.5\\
                OVRL-v2-IIN~\cite{yadav2023ovrl} & \ding{55} & \ding{51} & \ding{55} & 24.8 & 11.8\\
                Mod-IIN~\cite{krantz2023navigatingobjectsspecifiedimages} & \ding{51} & \ding{51} & \ding{55} & 56.1 & 23.3\\
                Goat~\cite{chang2024goatthing} & \ding{51} & \ding{55} & \ding{51} & 50.6 & 24.1 \\
                UniGoal~\cite{yin2025unigoaluniversalzeroshotgoaloriented} & \ding{51} & \ding{55} & \ding{51} & 60.2 & 23.7\\
                \midrule
                PSL~\cite{sun2024prioritizedsemanticlearningzeroshot} & \ding{51} & \ding{51} & \ding{51} & 23.0 & 11.4\\
                \rowcolor{gray!30}\textbf{ImagineNav++} & \ding{51} & \ding{51} & \ding{51} & 52.4 & 32.8 \\
                \rowcolor{gray!30}\textbf{ImagineNav++(Oracle)} & \ding{51} & \ding{51} & \ding{51} & 55.6 & 37.1 \\
                \bottomrule
            \end{tabular}
        }}
    \end{minipage}
\end{table*}

\begin{table*}[!t]
    \centering
    \small
 \renewcommand{\arraystretch}{1.2}
    \renewcommand{\tabcolsep}{5pt}    
    \setlength{\abovecaptionskip}{-0.0cm}   \setlength{\belowcaptionskip}{-0.0cm}    
 \caption{\textbf{ImagineNav++: ablation study on the imagination and memory modules}. `Imagination' refers to whether the future imaginations are used as visual prompts of the VLM. When it is removed, we feed current observations into VLM for deciding the best exploration direction, and set the next waypoint 2 meters away from the current location along the direction. Here, the distance of 2 meters is considered as it is comparable to that generated by T=11. `NVS' indicates whether the image is captured from a real environment or synthesized via the NVS model. }
 \vspace{0.1cm}
 \label{table2}
        \small
        \resizebox{0.72\textwidth}{!}{
        \begin{tabular}{ccccccccccc}
            \toprule
              \multirow{2}{*}{\textbf{Imagination}} & \multirow{2}{*}{\textbf{Where2Imagine}} & \multirow{2}{*}{\textbf{NVS}} & \multirow{2}{*}{\textbf{Memory}} & \multicolumn{2}{c}{\textbf{ObjectNav}} & \multicolumn{2}{c}{\textbf{InsINav}} \\
                \cmidrule{5-6} \cmidrule{7-8}
                 & & & & SR$\uparrow$ & SPL$\uparrow$ & SR$\uparrow$ & SPL$\uparrow$\\
                \midrule
                \ding{55} & \ding{55} & Oracle & \ding{55} & 43.0 & 24.7 & 34.9 & 24.1\\
                \ding{51} & \ding{55} & Oracle & \ding{55} & 55.0 & 27.6 & 47.2 & 34.4\\
                \ding{51} & \ding{51} & Oracle & \ding{55} & 64.0 & 28.3 & 50.3 & 36.7\\
                \ding{51} & \ding{51} & Oracle & \ding{51} & 70.0 & 36.9 & 51.9 & 38.2\\
                \ding{51} & \ding{55} & PolyOculus & \ding{55} & 49.0 & 23.3 & 37.2 & 23.0 \\
                \ding{51} & \ding{51} & PolyOculus & \ding{55} & 56.0 & 24.3  & 43.4 & 30.3\\
                \ding{51} & \ding{51} & PolyOculus & \ding{51} & 67.0 & 30.4 & 48.8  & 33.1 \\
            \bottomrule
        \end{tabular}
        }
        \vspace{-0.3cm}
\end{table*}
\subsection{Numerical Evaluations on \texttt{InsINav} Task}
Table~\ref{table:instance} summarizes the performance of our ImagineNav++ and baseline methods on the InsINav task. While UniGoal~\cite{yin2025unigoaluniversalzeroshotgoaloriented} achieves the highest SR of 60.2\%, it requires a graph-based planner that introduces significant computational overhead from graph construction, graph update, and graph matching. In comparison, our ImagineNav++ operates under a strict mapless setting and still attains a competitive SR of 52.4\%. More notably, ImagineNav++ establishes a new state-of-the-art in path efficiency and achieves a superior SPL of 32.8\%, significantly outperforming all other methods, including those that leverage map information (Mod-IIN~\cite{krantz2023navigatingobjectsspecifiedimages}: 23.3\%, Goat~\cite{chang2024goatthing}: 24.1\%, UniGoal~\cite{yin2025unigoaluniversalzeroshotgoaloriented}: 23.7\%). This advantage stems from our proactive planning paradigm, wherein the agent utilizes an imagination mechanism to synthesize and evaluate potential gains across multiple future directions, and subsequently integrates historical observations to select the optimal exploratory path, thereby minimizing haphazard exploration and redundant detours. This demonstrates that our ImagineNav++ achieves a superior navigation efficiency, excelling in both final goal attainment and the optimality of the paths taken. Among methods adhering to the same open-vocabulary, mapless, and universal constraints, our ImagineNav++ delivers markedly better performance than PSL~\cite{sun2024prioritizedsemanticlearningzeroshot}, more than doubling both SR (52.4\% vs. 23.0\%) and SPL (32.8\% vs. 11.4\%). These results highlight the capability of our ImagineNav++ to enable efficient and robust navigation relying solely on visual observations, without requiring explicit map construction. The efficacy of our framework on instance‑level objectives further validates the strong adaptability of our ImagineNav++ across diverse goal‑oriented visual navigation tasks.

\subsection{Ablation Study on Main Components}
\begin{figure}[H]
    \includegraphics[width=0.5\textwidth]{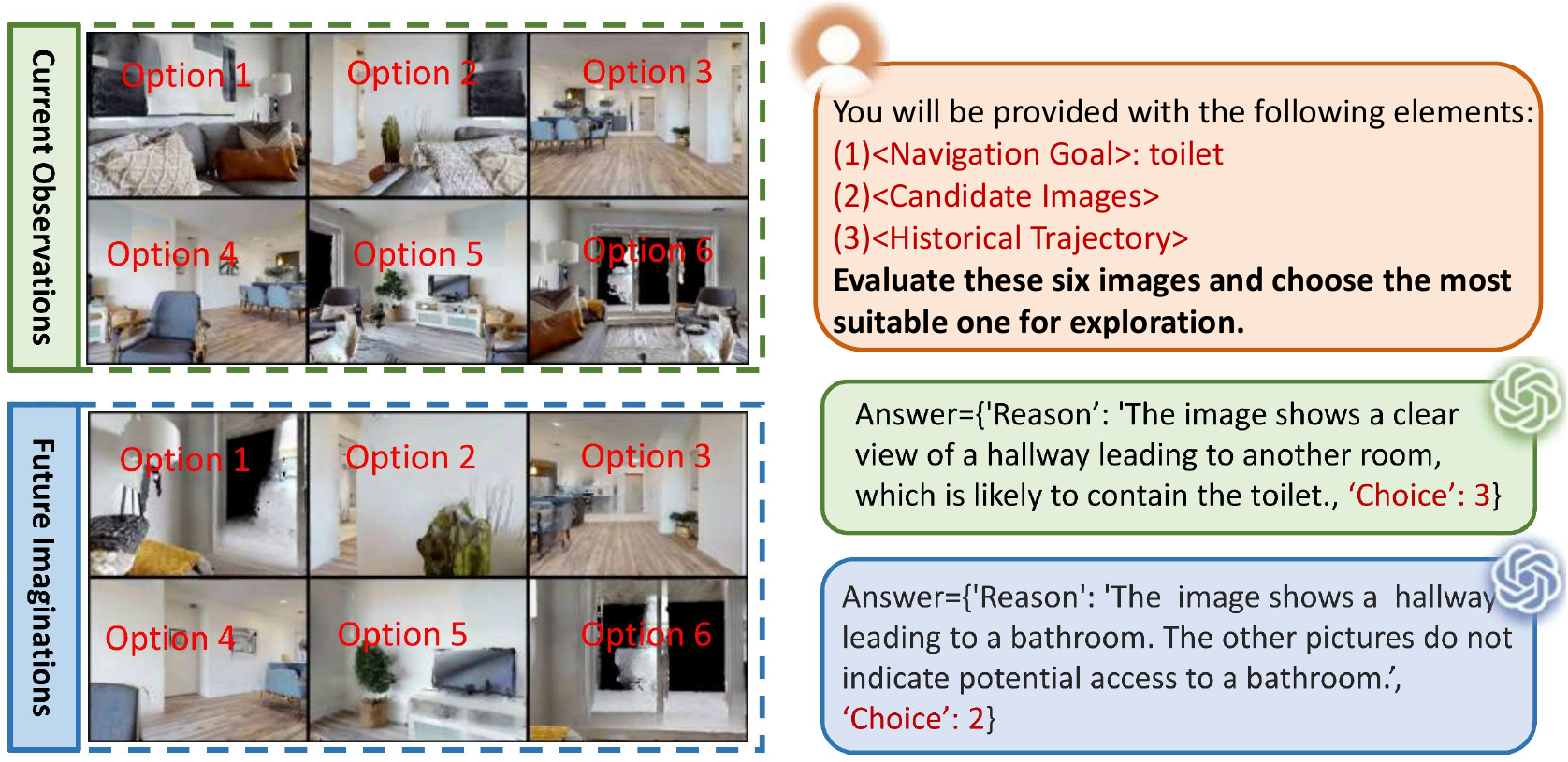}
   \caption{Visualization of future navigation waypoints generated by the imagination module shows significant semantic variation compared to the consistent semantics of current observations. This diversity demonstrates the module’s effectiveness in enhancing VLM decision-making.}
    \label{fig4}
\end{figure}

We conducted an ablation study to systematically evaluate the contribution of key components in the proposed method: \textit{Imagination}, \textit{Where2Imagine}, and \textit{Memory}, with each variant evaluated for 100 epochs. As shown in Table~\ref{table2}, introducing  Imagination (Row~\#2) increases SR by significant margins of 12.0\% (55.0\% vs. 43.0\%) on ObjectNav and 12.3\% (47.2\% vs. 34.9\%) on InsINav by utilizing future-view imaginations as visual prompt of the VLM for deciding exploration direction. Please note that in the configuration of Row~\#2 (without Where2Imagine), future views are generated at six locations, each positioned two meters from the agent along the directions of its current observable viewpoints. Further addition of the Where2Imagine module (Row~\#3 and \#6) boosts ObjectNav SR by 9.0\% (`w/o NVS') and 7.0\% (`NVS') while improving InsINav SR by 3.1\% and 6.2\%, respectively. The observed performance gain demonstrates the critical role of future imagination in enhancing the VLM's spatial reasoning capability. Such improvements can be attributed to the greater semantic disparity between different imaginations, as illustrated in Figure~\ref{fig4}. Finally, incorporating the Memory mechanism (Row~\#4 and \#7) leads to the highest performance across all metrics, notably improving ObjectNav SR by 11.0\% and InsINav SR by 5.4\% under the setting of `NVS', underscoring its value in maintaining long-term spatial reasoning. In particular, as previously discussed, the performance of ImagineNav++ is constrained by the quality of the off-the-shelf NVS model~\cite{yu2024polyoculussimultaneousmultiviewimagebased}, as reflected in the performance gaps between Rows~\#3 and \#6 and between Rows~\#4 and \#7.
The proposed Memory module, however, effectively mitigates this dependency and substantially alleviates the performance degradation caused by imperfect NVS synthesis. This is demonstrated by the relatively smaller performance drops on ObjectNav (64.0$\rightarrow$56.0 vs. 70.0$\rightarrow$67.0) and on InsINav (50.3$\rightarrow$43.4 vs. 51.9$\rightarrow$ 48.8). This is reasonable, as the Memory module allows the agent to accumulate historical observations, thereby building a more consistent and reliable internal representation of the environment. This process, in turn, reduces the agent's reliance on any single, and potentially flawed, NVS synthesis when making navigation decisions.
\subsection{Analysis of Where2Imagine Module}
To validate the effectiveness of the proposed design, we perform ablation studies under a memory-free configuration across all experiments reported in this section.
\subsubsection{\textbf{Choice of Imagination Horizon}} 
\begin{figure}[H]  
    \includegraphics[width=0.5\textwidth,height=1.8in]{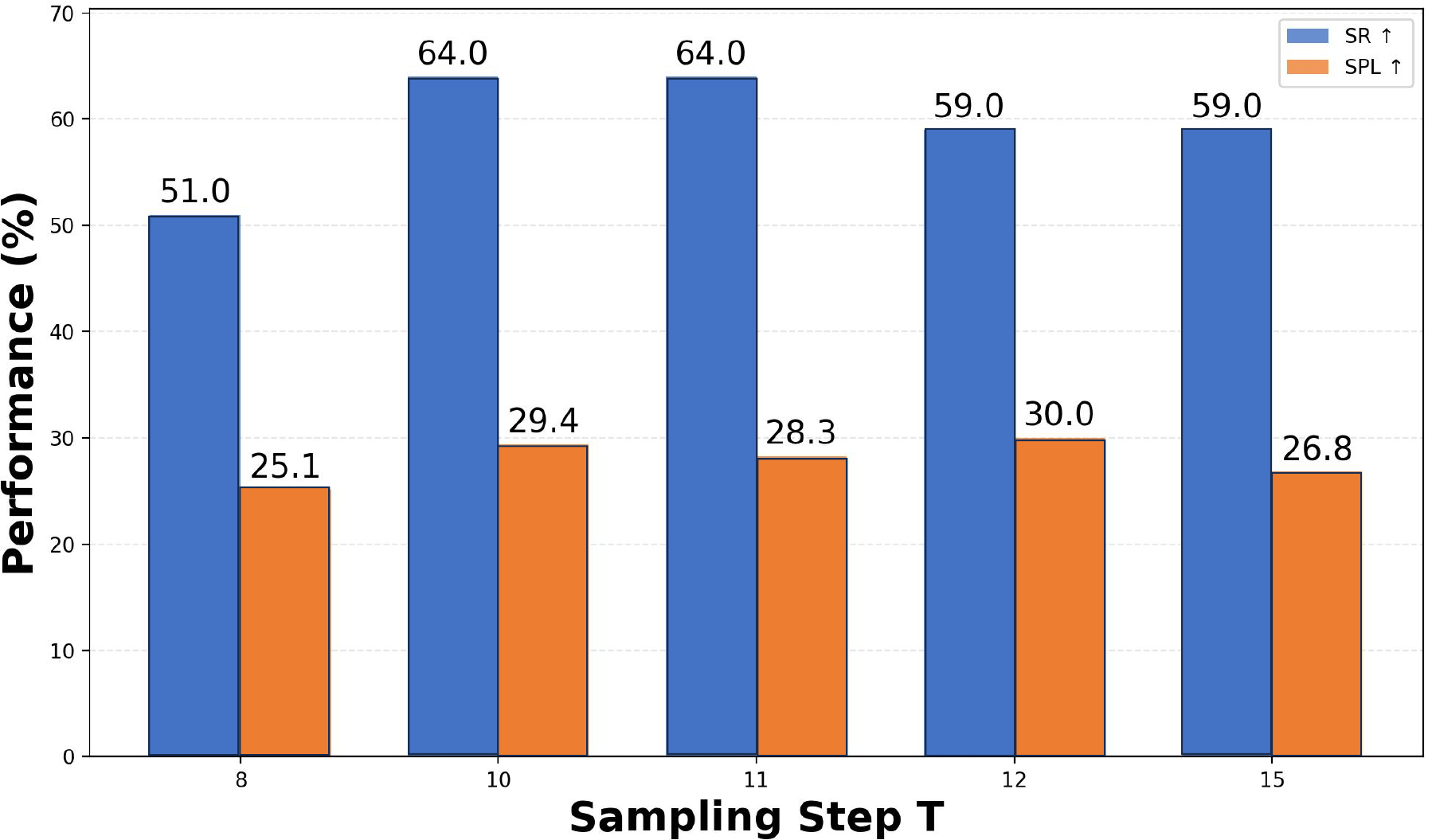}
    \caption{\textbf{Where2Imagine:} the impact of different sampling intervals $\mathbf{T}$ on ObjectNav performance on HM3D dataset.}
    \label{sample_t}
\end{figure}

\begin{figure}[H]  
    \includegraphics[width=0.5\textwidth]{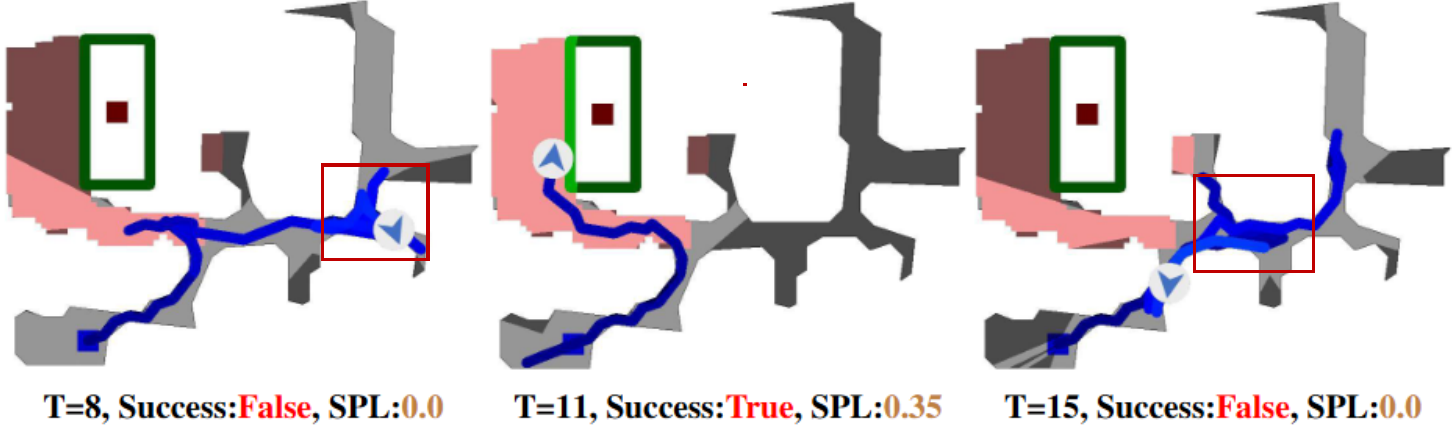}
    \caption{Comparison of trajectories at different sampling steps T. This image presents a top-down view of the entire trajectory as the agent searches for the target (a chair). The red box highlights the situation where the agent encounters a local trap during navigation.}
    \label{fig5}
\end{figure}

We explore the impact of the sampling step $T$ on the final navigation performance by varying $T$ from 8 to 15. For each $T$, we re-generate the labeled image data and re-train the ResNet-18 for relative pose prediction. Each variant was tested for 100 epochs under conditions where the agent had access to real panoramic observations. As shown in Figure~\ref{sample_t}, the best success rate and SPL are obtained when $T$ is set to 11. Furthermore, we visualize several navigational trajectories under different values of $T$ in Figure~\ref{fig5} to facilitate explanation. As can be seen,  when $T$ is relatively small (i,e., 8), the agent is easily trapped as marked by red square, since it mainly resorts to local semantic information for inferring its exploration direction, making it susceptible to converging on suboptimal local solutions. Conversely, when $T$ is excessively large, although the agent has access to more distant information, it is prone to miss some critical intermediate semantics which are closely related to target and are worth exploring, leading to erroneous long-range decisions, particularly in intricate environments. However, an optimally calibrated $T$ can strike a delicate balance between exploration and perception, thereby facilitating to obtain impressive navigation performance.
\vspace{-0.2cm}
\begin{figure*}[ht]
\centering
\includegraphics[width=1.00\linewidth]{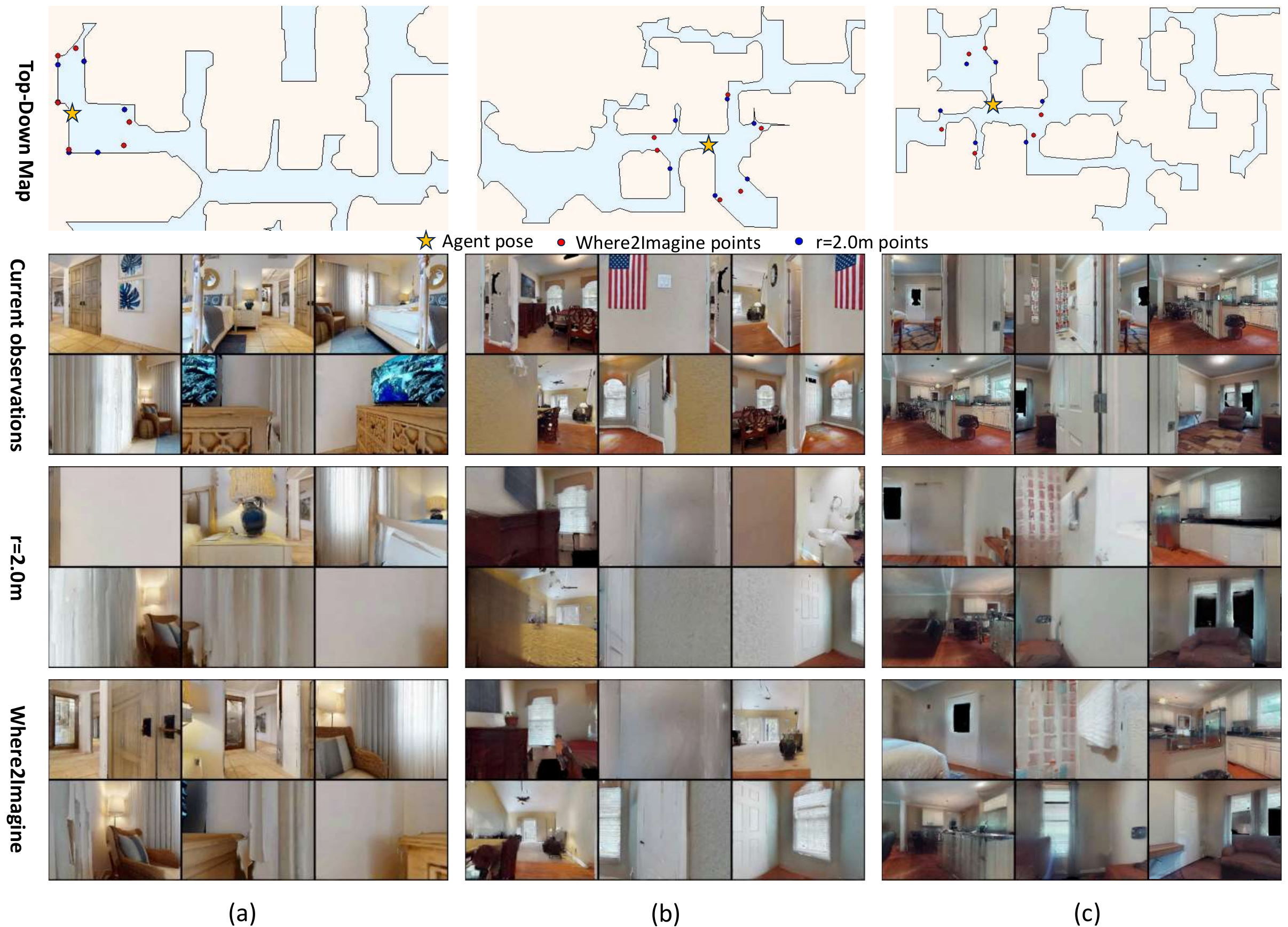}
\caption{The visualization of the relative pose predicted by our Where2Imagine module and uniform sampling~(radius: 2.0m; angular interval: $60^\circ$). The upper part shows the agent's current position (star marker) in different environments, as well as the distribution of the relative poses predicted by the Where2Imagine module (red dots) and uniform sampling (blue dots). The lower part shows the first-person views from different poses. Compared to uniform sampling, our Where2Imagine module tends to predict more towards walkable areas and directions with higher information density.}
\label{fig7}
\end{figure*}
\subsubsection{\textbf{Choice of Backbone.}} 
We evaluate the impact of different visual backbones on both waypoint prediction and final navigation performance. All backbones were adapted for our task: ResNet-18 and ViT were modified at their final layers and trained end-to-end, whereas DINOv2 and MAE were kept frozen, with their features fed into a separate trainable five-layer MLP. All experiments used real RGB observations without resorting to the NVS model and were tested for 100 epochs. As shown in Table~\ref{table4}, ResNet-18 achieves the best performance in both waypoint prediction and ObjectNav, despite being the most lightweight architecture. Furthermore, the significant performance gap across different backbones underscores the importance of our Where2Imagine module, indicating the usefulness of learning from human demonstrations. We also note a slight discrepancy between the waypoint prediction loss and the navigation success rate for DINOv2 and MAE. This is likely attributable to the domain shift between the MP3D dataset (used for waypoint evaluation) and the HM3D environment (used for navigation), which introduces evaluation variability.
\begin{table}[H]     
        \centering 
        \makeatletter\def\@captype{table}
        \caption{\textbf{Where2Imagine:} the impact of different \textbf{backbones} on ObjectNav performance. Loss refers to the test loss of Where2Imagine. TFS: training from scratch, FT: fine-tuning.}
        {\small
        \begin{tabular}{cccccccccc}
            \toprule
            \multirow{2}{*}{\textbf{Backbone}} & \multirow{2}{*}{\textbf{Params}} & \multirow{2}{*}{\textbf{Flops}} & \multirow{2}{*}{\textbf{Loss}} & \multicolumn{2}{c}{\textbf{HM3D}} \\
            \cmidrule{5-6}
            & & & & SR $\uparrow$ & SPL $\uparrow$ \\
            \midrule
            \textbf{ResNet-18 (TFS)} &11.4M &1.8G &0.12 &64.0  &28.3 \\
            \textbf{ResNet-18 (FT)} &11.4M &1.8G &0.24 &61.0  &29.7 \\
            \textbf{ViT (TFS)} &86.0M &16.9G &0.22 &61.0  &29.7 \\
            \textbf{ViT (FT)} &86.0M &16.9G &0.23 &58.0  &31.0 \\
            \textbf{DINOv2} &22.6M &5.5G &0.22 &58.0  &27.9 \\
            \textbf{MAE} &87.1M &4.4G &0.20 &57.0  &26.5 \\
            \bottomrule
        \end{tabular}
        }
        \label{table4}
\end{table}

\subsubsection{\textbf{Where2Imagine vs. Uniform Sampling for Exploration}}
Figure~\ref{fig7} presents a comprehensive qualitative evaluation of our proposed Where2Imagine module on the HM3D dataset. This visualization reveals several key insights. First, the waypoints generated by our Where2Imagine module are concentrated in navigable regions, yielding observation views that are semantically richer and capture essential structural elements such as furniture, doors, and hallways. This stands in clear contrast to the uniform sampling strategy (with a fixed radius of 2.0m and angular interval of $60^\circ$), which disregards scene geometry and frequently fails to capture critical semantic information. This observation confirms the capability of our Where2Imagine module for semantic-aware planning, aligning its exploratory behavior with perceptually salient structural cues. Another key finding emerges from comparing the ground-truth and synthesized view observations. Specifically, although the NVS model is prone to introducing artifacts and inaccuracies, notably in the synthesis of fine details, they generally achieve fundamental consistency with the true scene observation in both semantics and overall structure across diverse HM3D environments. This high-level consistency enables effective spatial reasoning in LLMs, which explains the performance gains observed in navigation tasks.

\subsection{Analysis of Selective Foveation Memory}

\begin{figure*}[ht]
\centering
\includegraphics[width=1.00\linewidth]{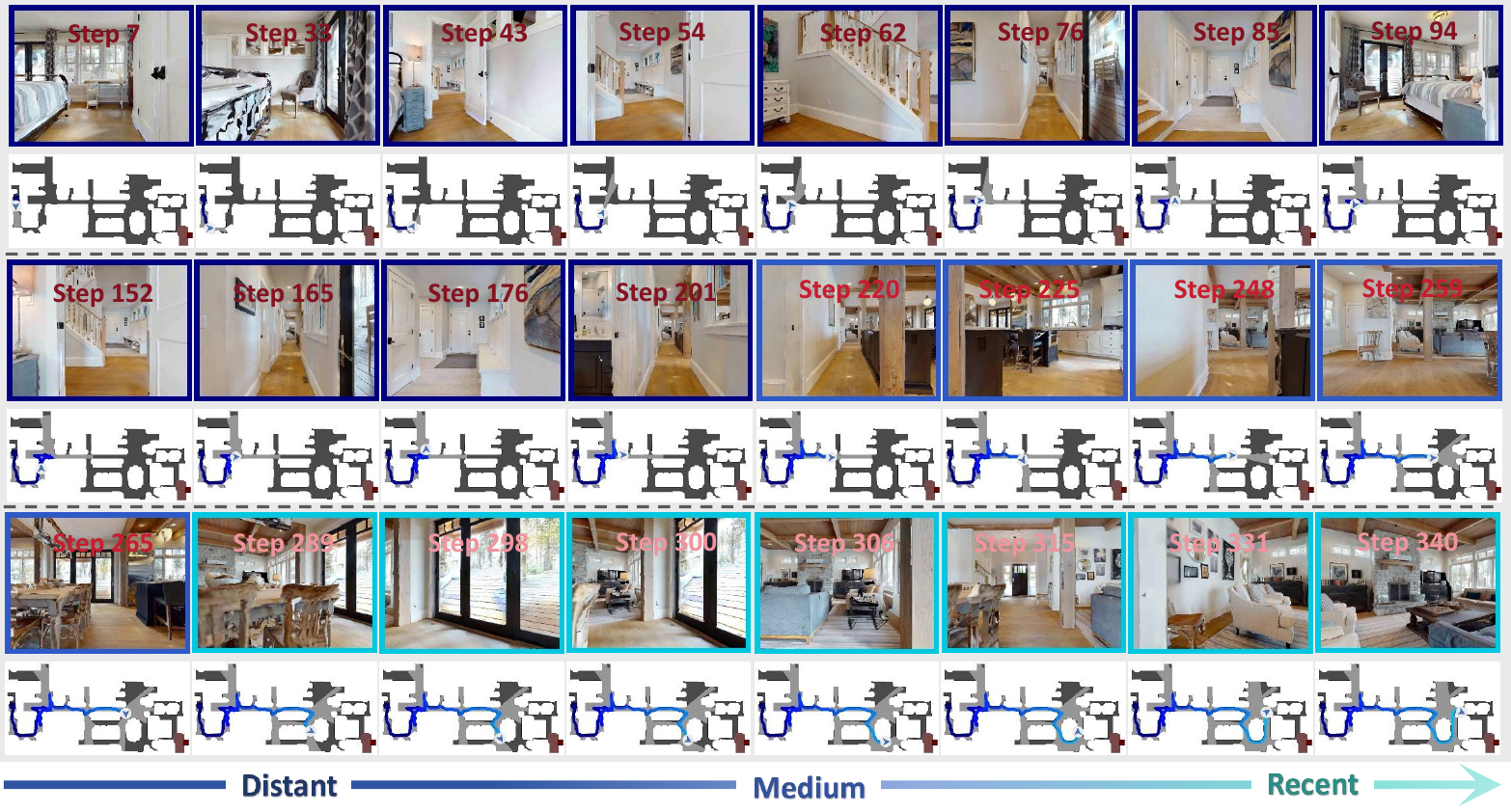}
\caption{Visualization of our hierarchical keyframe-based memory, depicting RGB observations and their corresponding top-down map positions for selected keyframes. Color intensity encodes the temporal stage of each memory entry, with darker hues representing earlier time steps. Results indicate that our memory scheme effectively prioritizes structurally critical locations, such as intersections and corners, thereby optimizing memory allocation through feature saliency.}
\label{fig:keyframe}
\end{figure*}

\begin{figure*}[ht]
\centering
\includegraphics[width=1.00\linewidth]{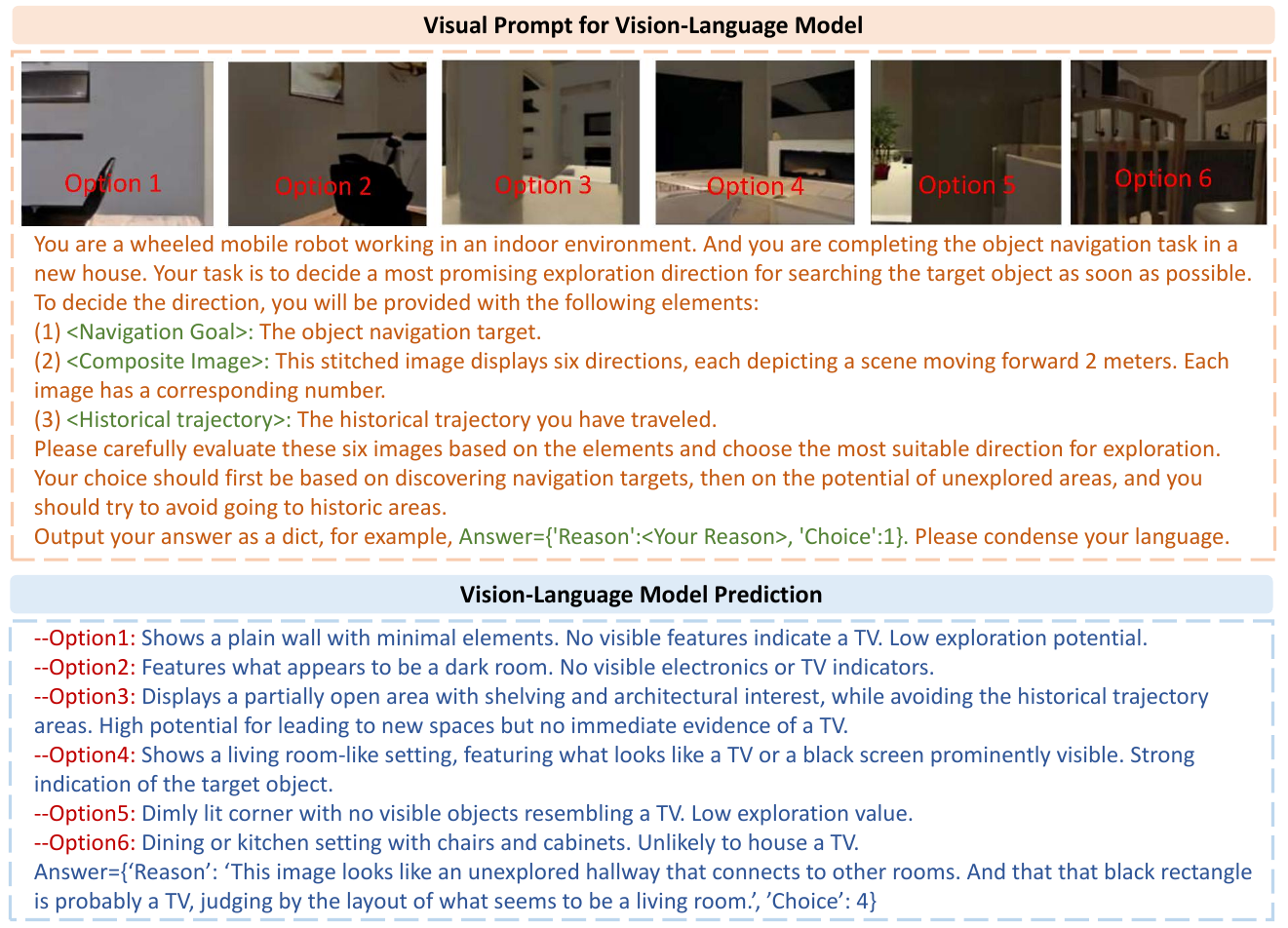}
\caption{Complete prompt input and decision output of the Vision-Language model for exploration direction selection in the object goal navigation task.}
\label{fig6}
\end{figure*}

We evaluate the impact of different memory configurations on final ObjectNav performance.
As presented in Table~\ref{table：memory}, the baseline configuration `w/o' (without memory) achieves a SR of 56.0 and SPL of 24.3. Interestingly, the 'Full' memory approach, which utilizes all historical observations, does not improve SR compared to the no-memory baseline, though it does yield a slight improvement in SPL (26.2 vs. 24.3), indicating marginally more efficient path planning. This could be explained by the fact that the full memory, while providing complete historical information, also introduces substantial redundancy and potential distraction, which hampers the identification of actionable navigation cues and thus fails to improve the final success rate. In comparison, the 'Uniform' keyframe-based memory, which applies the same keyframe selection threshold (0.8) across all temporal horizons, brings a notable performance gain, achieving a SR of 65.0 and SPL of 28.2. This result confirms the critical role of perceptually salient landmarks in visual navigation, demonstrating the efficacy of our keyframe-based memory mechanism in robustly identifying and retaining such structural cues. Most importantly, our proposed `Selective' memory mechanism, which employs varying keyframe thresholds (0.8 for recent, 0.73 for medium-term, and 0.6 for long-term history), achieves the best results, i.e., a SR of 67.0 and SPL of 30.4. This demonstrates that a non-uniform keyframe selection strategy, favoring more frequent keyframes in recent history and sparser ones in the distant past, better supports both navigation success and path efficiency while improving inference efficiency by reducing the amount of keyframes compared to the uniform keyframe memory mechanism. Furthermore, we also evaluate the real-time incremental update variant. This streaming approach drastically reduces per-step memory update latency from 1.70\,s to 0.16\,s (\(>10\times\) speedup). Inevitably, the hierarchical ``keyframe of keyframes'' approximation introduces minor information quantization error compared to the global optimal variant, yielding an SR of 64.0 and SPL of 28.1--a marginal performance trade-off. With strict $\mathcal{O}(1)$ bounds on computation and storage, this variant represents a highly efficient and practical solution for continuous real-world deployment.

\begin{table}      
        \centering
        \makeatletter\def\@captype{table}
        \caption{The Impact of Different Memory Configurations on ObjectNav Performance.}
        {\small
        \setlength{\tabcolsep}{3pt}
        \begin{tabular}{cccccccccc}
            \toprule
            \multirow{2}{*}{\textbf{Memory}} & \multirow{2}{*}{\textbf{Sim. Thresh.}} & \multirow{2}{*}{\textbf{Avg. Mem.}} & \multicolumn{2}{c}{\textbf{HM3D}} \\
            \cmidrule{4-5}
            & & & SR $\uparrow$ & SPL $\uparrow$ \\
            \midrule
            \textbf{w/o} &- &- &56.0  &24.3 \\
            \textbf{Full} &- &122.3 &56.0  &26.2 \\
            \textbf{Uniform} &0.8 \textbf{/} 0.8 \textbf{/} 0.8 &58.3 &65.0  &28.2 \\
            \textbf{Selective (Global)} & 0.8 \textbf{/} 0.73 \textbf{/} 0.6 &21.8 &67.0  &30.4 \\
            \textbf{Selective (Incremental)} & 0.8 \textbf{/} 0.73 \textbf{/} 0.6 &21.8 &64.0  &28.1 \\
            \bottomrule
        \end{tabular}
        }
        \label{table：memory}
        \vspace{-0.5cm}
\end{table}

Figure~\ref{fig:keyframe} presents exemplary keyframes selected during navigation. The results demonstrate that our strategy constructs a compact environmental representation by selecting distinctive structural features, such as living room layouts and corridor characteristics. This enables the agent to form and maintain a comprehensive and efficient spatial awareness. Importantly, as shown in the figure, the method allows the agent to recognize previously explored areas, effectively avoiding local hesitation or cyclic paths, which validates the efficacy of the keyframe extraction algorithm in improving navigation efficiency. It is also noteworthy that for long-horizon navigation (up to 500 steps), our method achieves compact environmental representation using only an average of about 20 keyframes, as shown in Col~\#3 of Table~\ref{table：memory}. This sparse representation significantly reduces the LLM’s input tokens, enhancing inference efficiency substantially without compromising navigation accuracy.

\begin{figure*}[ht]
    \includegraphics[width=0.95\textwidth]{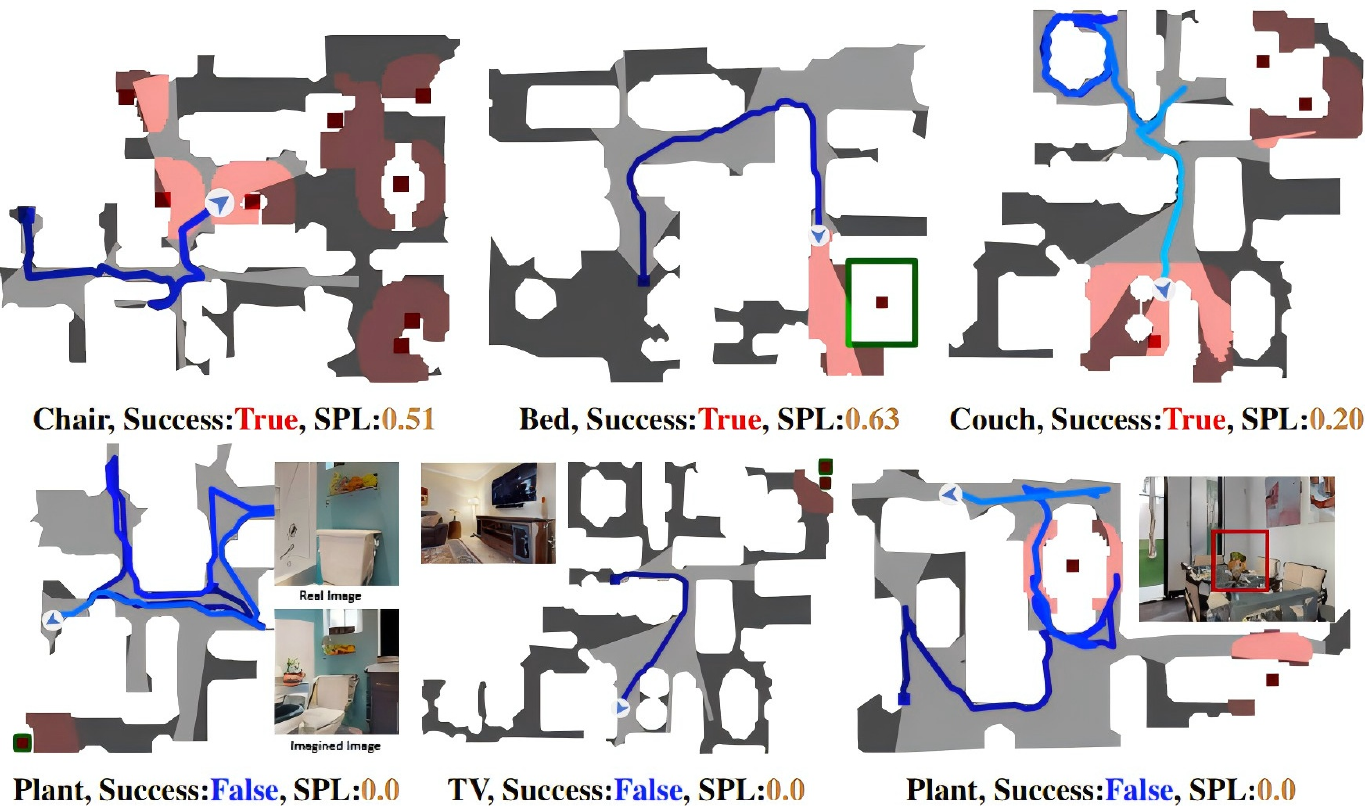}
    \caption{Analysis of object-goal navigation trajectories: success vs. failure examples. The top and bottom rows compare top-down paths from successful and unsuccessful episodes, identifying critical factors for task completion.}
    \label{fig4-1}
\end{figure*}
\subsection{Analysis of VLM Planner}
 We conducted a comparative evaluation of the effects of different VLMs on navigation performance, as detailed in Table \ref{table5}. The experiments used real RGB without NVS model. The results demonstrate that GPT-4o-mini~\cite{openai2024gpt4omini} and GPT-4-Turbo~\cite{openai2023gpt4turbo} exhibit negligible differences in success rate and SPL metrics, while substantially outperforming LLaVA~\cite{10655294}, demonstrating that advanced reasoning capabilities are crucial for navigation tasks. Furthermore, our results indicate that within the same model family, strategically selecting more cost-effective variants (such as GPT-4o-mini vs. GPT-4-Turbo) achieves comparable navigation performance while dramatically reducing operational costs by approximately 98\%. This optimization enables more efficient resource allocation and faster inference without sacrificing navigation performance, providing practical guidance for real-world deployment scenarios where computational efficiency and operational costs are significant concerns. 
 \begin{figure}[ht]  
    \includegraphics[width=0.45\textwidth]{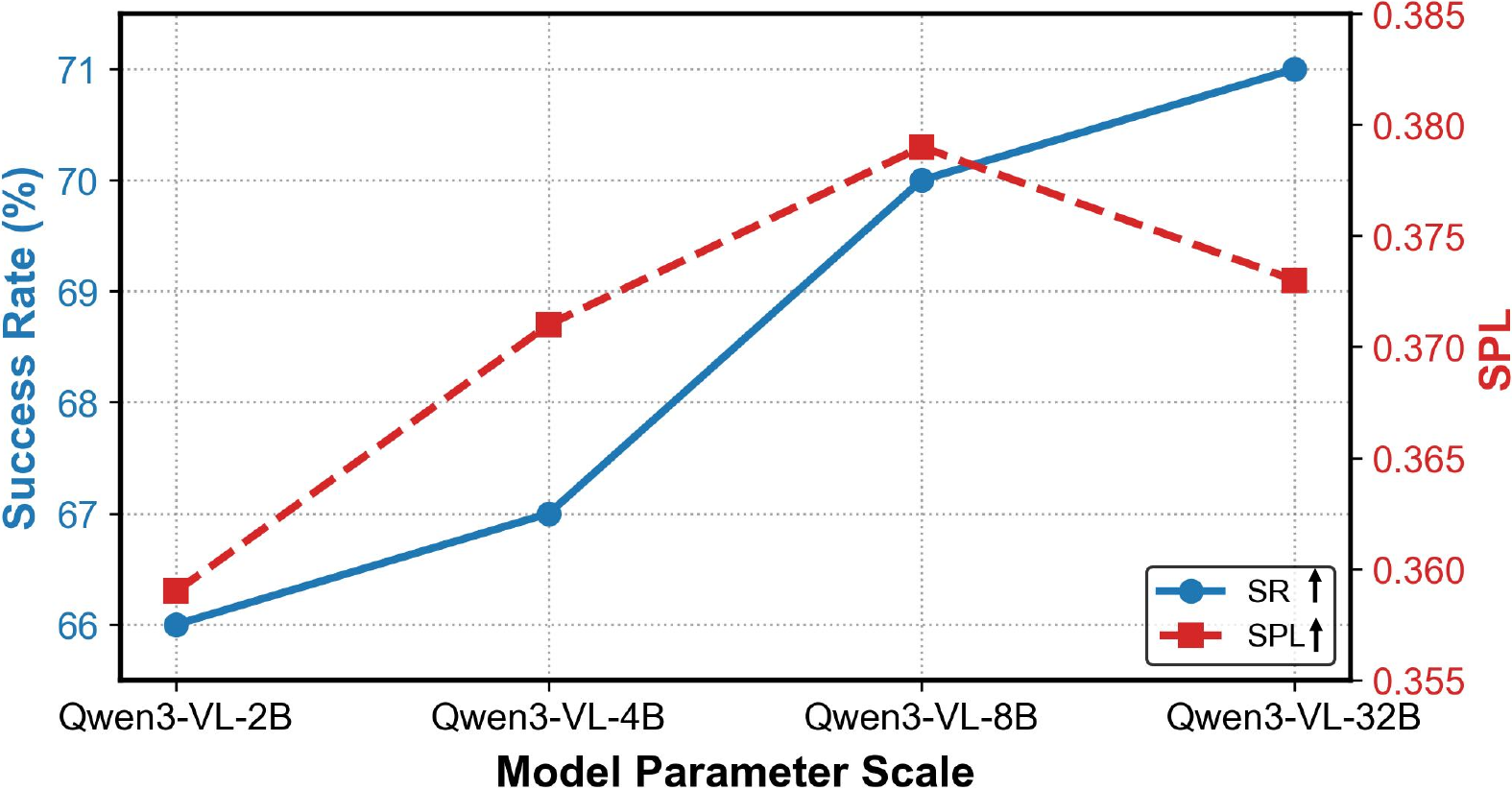}
    \caption{Scaling Study of Open-source VLMs for ObjectNav.}
    \label{scaling_study}
\end{figure}

 \begin{table}[ht]
 \vspace{-0.1cm}
    \centering
    \footnotesize
    \renewcommand{\arraystretch}{0.9}
    \renewcommand{\tabcolsep}{4pt}    
 \caption{The effect of different VLMs on ObjectNav Performance.}
 \label{table5}
        \small
          \begin{tabular}{ccccccccc}
            \toprule
              \multirow{2}{*}{\textbf{VLM}} & \multirow{2}{*}{\textbf{Cost (per Mtok.)}}& \multicolumn{2}{c}{\textbf{HM3D}} \\
                \cmidrule{3-4}
                 & & SR $\uparrow$ & SPL $\uparrow$ \\
                \midrule
                \textbf{LLaVa}~\cite{10655294} & N/A &48.0  &25.4 \\
                \textbf{GPT-4-Turbo}~\cite{openai2023gpt4turbo} & \$20.00 &66.0  &36.3 \\
                \textbf{GPT-4o-mini}~\cite{openai2024gpt4omini} & \$0.375 &70.0  &36.9 \\
                \textbf{Qwen3-vl-32b}~\cite{bai2025qwen3vltechnicalreport} & N/A &71.0  &37.3 \\
                \textbf{Qwen3.5-27b}~\cite{qwen3.5} & N/A &71.0  &40.7 \\
            \bottomrule
        \end{tabular}
\end{table}

To examine the dependence on closed-source APIs and assess edge-deployment feasibility, we further evaluated ImagineNav++ using the state-of-the-art open-source Qwen series~\cite{bai2025qwen3vltechnicalreport,qwen3.5}. As reported in Table~\ref{table5}, locally deployed models such as Qwen3-VL-32B~\cite{bai2025qwen3vltechnicalreport} and Qwen3.5-27B~\cite{qwen3.5} surprisingly achieve SRs of 71.0\% with corresponding SPLs of 37.3 and 40.7, slightly surpassing the GPT-4o-mini baseline. These results indicate that ImagineNav++ maintains strong performance without reliance on proprietary cloud VLMs, highlighting both the framework’s robustness and its practical applicability for edge-computing scenarios.

Furthermore, we conducted a systematic scaling study across different parameter sizes of the Qwen3-VL family (ranging from 2B to 32B). As illustrated in Figure~\ref{scaling_study}, the results reveal a clear scaling trend: SR generally increases with model size, while SPL improves up to Qwen3-VL-8B and exhibits a marginal decrease for the largest model, suggesting slight saturation. Notably, even the relatively lightweight models (e.g., Qwen3-VL-8B) achieve a highly competitive SR of 70.0\% while demanding substantially lower computational resources. This consistent scaling behavior underscores the strong potential of deploying our mapless ImagineNav++ entirely offline on resource-constrained robotic platforms, demonstrating its practicality for edge-computing scenarios.

\vspace{-0.1cm}
\subsection{Decision-Making Details of VLM Planner}
The input data for the VLM is constructed as follows. Based on the agent’s current pose, an RGB observation is captured. The Where2Imagine module then predicts a relative pose from this observation. Both the RGB observation and the predicted relative pose are fed into the NVS to synthesize a novel view image. This procedure is repeated every 60°, generating a total of six novel view images. These images are stitched together and combined with historical context to form the complete input to the VLM. In the ablation studies, the VLM inputs vary depending on whether Where2Imagine is used for relative pose prediction and novel view synthesis; however, in all cases, the initial input at the beginning of each cycle is the RGB observation captured from the agent’s position at 60° intervals. The full prompts and VLM responses are provided in Figure~\ref{fig6}.

\subsection{Analysis of Successful and Failed Trajectories}
As shown in Figure~\ref{fig4-1}, our method enables efficient path planning and navigation toward diverse targets. In particular, the top-middle subfigure illustrates a scenario where the agent must traverse multiple rooms. Although such complex environments increase the risk of disorientation, our ImagineNav++ successfully reaches the goal after exploration, demonstrating its robustness in long-horizon, multi-room settings. Furthermore, as depicted in the upper-right subfigure, the memory module helps detect local search loops from historical observations and autonomously adjust the exploration direction. We also present some failure examples at the bottom of Figure~\ref{fig4-1}. We identified three key factors contributing to these navigation failures. First, the synthesized image from the NVS does not align with the real observation, such as creating objects that are not present in the real scene as shown in the bottom left of Figure~\ref{fig4-1}, which causes the VLM to make incorrect inference. Second, some object instances are neglected for marking by the simulator, and therefore a successful trajectory is wrongly considered as a failure (a.k.a. false failure) as shown in the middle of Figure~\ref{fig4-1}. Moreover, the visual ambiguity of the target leads directly to recognition failure, causing the agent to terminate the navigation task unsuccessfully due to the lack of a key basis for the stopping decision, as shown in the bottom right of Figure~\ref{fig4-1}.

\section{Conclusion}

In this work, we present ImagineNav++, a novel framework that successfully repurposes pre-trained vision-language models as efficient embodied navigators for mapless, open-vocabulary visual navigation. Our approach leverages a learned future-view imagination module to generate semantically meaningful candidate observations, which then serve as direct visual prompts for the VLM to identify the most informative next viewpoint. Augmented with a selective foveation memory that hierarchically preserves essential spatial and semantic information from historical observations, the framework sustains spatial consistency and facilitates robust temporal reasoning without requiring fine-tuning. Extensive experiments confirm its state-of-the-art performance in terms of both effectiveness and efficiency. Future work may explore extending the framework to support multimodal goal specifications (i.e., text-goal navigation), and reducing inference latency to enable real-time operation on resource-constrained robotic platforms. Further investigation into integrating lifelong adaptation mechanisms could also enhance the system’s ability to continuously evolve in dynamic environments, opening avenues for sustained autonomous operation.

\section*{Acknowledgments}
This work was in part supported by the National Natural Science Foundation of China (Grant No. 62273093).

\bibliographystyle{IEEEtran}
\bibliography{ref}

\vfill

\end{document}